\documentclass[11pt]{article}
\usepackage[margin=1in]{geometry}

\usepackage{amsmath,amsthm,amssymb,amsfonts,cancel}
\usepackage{thmtools,enumitem,subfigure}
\usepackage{algcompatible}
\usepackage{algorithm}
\usepackage[noend]{algpseudocode}
\usepackage{mathtools}
\usepackage{dsfont}
\usepackage{tcolorbox}
\usepackage{bbm}
\usepackage{todonotes}
\usepackage{tensor}
\usepackage{caption}
\usepackage{bm}
\usepackage{caption}
\usepackage{hyperref}
\usepackage{placeins}
\usepackage{authblk}
\usepackage{epsfig}
\usepackage{natbib}
\usepackage[utf8]{inputenc}
\usepackage[T1]{fontenc}
\usepackage[stable]{footmisc}

\usepackage{todonotes}
\usepackage{textcomp}

\renewcommand{\S}{\mathcal{S}}
\renewcommand{\P}{\mathcal{P}}
\newcommand{\D}{\mathcal{D}}
\newcommand{\A}{\mathcal{A}}

\newcommand{\dom}[1]{\texttt{dom}\,\left( #1 \right)}
\newcommand{\Ind}[1]{\mathds{1}_{\left[ #1 \right]}}

\newcommand{\Exp}[1]{\mathbb E \left[ #1 \right]}

\renewcommand{\Pr}{\mathbb{P}}
\newcommand{\pfail}{\delta}
\newcommand{\Qhat}[2]{\mathbf{Q}_{#1}^{#2}}
\newcommand{\Vhat}[2]{\mathbf{V}_{#1}^{#2}}
\newcommand{\Bhat}{\mathbf{B}}

\newcommand{\taumin}{\tau_{\text{min}}}

\newcommand{\gym}{{\tt OpenAI Gym}}

\newcommand{\pendulum}{{\tt Pendulum}}
\newcommand{\cartpole}{{\tt CartPole}}

\PassOptionsToPackage{normalem}{ulem}
\usepackage{ulem}

\providecolor{added}{rgb}{0,0,1}
\providecolor{deleted}{rgb}{1,0,0}
\providecolor{addedtwo}{rgb}{0,0.5,0}
\mathchardef\mhyphen="2D % Define a "math hyphen"
\DeclareMathOperator*{\argmax}{\arg\!\max}
\DeclareMathOperator*{\dsmax}{\max}

\let\originalleft\left
\let\originalright\right
\renewcommand{\left}{\mathopen{}\mathclose\bgroup\originalleft}
\renewcommand{\right}{\aftergroup\egroup\originalright}
\newcommand\blfootnote[1]{%
  \begingroup
  \renewcommand\thefootnote{}\footnote{#1}%
  \addtocounter{footnote}{-1}%
  \endgroup
}

\newtheorem{theorem}{Theorem}
\numberwithin{theorem}{section}
\newtheorem{definition}[theorem]{Definition}

\makeatletter
\let\@fnsymbol\@arabic
\makeatother

\begin{document}

\title{Control with adaptive Q-learning}
\author{João P. Araújo\thanks{Email: \texttt{joao.p.araujo@tecnico.ulisboa.pt}} \qquad Mário A. T. Figueiredo$^2$
\qquad Miguel Ayala Botto$^3$
}
\date{\small{$^1$Instituto Superior Técnico, Universidade de Lisboa \\
$^2$Instituto de Telecomunicações; Instituto Superior Técnico, Universidade de Lisboa, Portugal.\\
$^3$IDMEC, Instituto Superior Técnico, Universidade de Lisboa, Portugal.}}
	\maketitle

	\begin{abstract}
This paper evaluates \textit{adaptive Q-learning} (AQL) and \textit{single-partition adaptive Q-learning} (SPAQL), two algorithms for efficient model-free episodic reinforcement learning (RL), in two classical control problems (\pendulum~and \cartpole). AQL adaptively partitions the state-action space of a \textit{Markov decision process} (MDP), while learning the control policy, \textit{i. e.}, the mapping from states to actions. The main difference between AQL and SPAQL is that the latter learns time-invariant policies, where the mapping from states to actions does not depend explicitly on the time step. This paper also proposes the SPAQL with terminal state (SPAQL-TS), an improved version of SPAQL tailored for the design of regulators for control problems. The time-invariant policies are shown to result in a better performance than the time-variant ones in both problems studied. These algorithms are particularly fitted to RL problems where the action space is finite, as is the case with the \cartpole~problem. SPAQL-TS solves the \gym~\cartpole~problem, while also displaying a higher sample efficiency than \textit{trust region policy optimization} (TRPO), a standard RL algorithm for solving control tasks.
Moreover, the policies learned by SPAQL are interpretable, while TRPO policies are typically encoded as neural networks, and therefore hard to interpret. Yielding interpretable policies while being sample-efficient are the major advantages of SPAQL.

\blfootnote{The code for the experiments is available at \url{https://github.com/jaraujo98/SinglePartitionAdaptiveQLearning}.}

\end{abstract}
\newpage
\setcounter{tocdepth}{2}
\tableofcontents
\newpage

\section{Introduction}

\textit{Reinforcement learning} (RL) is an area within machine learning that studies how agents (such as humanoid robots, self-driving cars, or computer programs that play chess) can learn how to perform their tasks without being explicitly told how to do so. The problem can be posed as that of learning a mapping from system states to agent actions. In order to do this, the agent observes and interacts with the system, choosing which action to perform given its current state. By choosing a certain action, the system transitions to a new state, and the agent may receive a reward or a penalty. Based on these rewards and penalties, the agent learns to assess the quality of actions. By setting the agent's objective to be reward maximization, it will learn to prefer good actions over bad ones.
As an example, consider a robot that stacks boxes. The system is the environment where the robot works, and the actions correspond to moving, grabbing, and releasing the boxes. The robot may receive rewards for each box successfully stacked, and may receive penalties for failing to stack a box, or knocking down the existing box stack.
The sum of all rewards and penalties received by the robot during its task is called the cumulative reward. Maximizing this cumulative reward is the objective of RL \citep{sutton2018}. This approach has led to many successful applications of RL in areas such as control of autonomous vehicles \citep{DLVehicles}, game playing \citep{Silver2018, openai2019dota}, healthcare \citep{Shortreed2011}, and education \citep{10.1145/3231644.3231672}, to name a few.

Existing RL methods can be divided into two main families: model-free and model-based. In model-based methods, the agent either has access to or learns a model of the system, which it uses to plan future actions. Although model-free methods are applicable to a wider array of problems, concerns with the empirical performance of model-free methods have been raised in the past due to their sample complexity \citep{pilco, trpo}. Two additional problems, besides sample complexity, are: the trade-off between exploration and exploitation, and dealing with \textit{Markov decision processes} (MDP) with continuous state and action spaces. Several solutions to deal with these two latter problems have been proposed. For example, exploration and exploitation can be balanced using a stochastic policy, such as $\varepsilon$-greedy \citep{sutton2018}. MDPs with continuous state and action spaces can be dealt with by using function approximators, such as neural networks.

Recently, some theoretical work has addressed the sample complexity of model-free methods. Informally, sample complexity can be defined as the number of samples that an algorithm requires in order to learn. Sample-efficient algorithms require fewer samples to learn, and thus are more desirable to develop and use. There are several approaches to sample-efficient RL \mbox{\citep{yu2018towards}}. However, until recently, it was not known whether model-free methods could be proved to be sample-efficient. This changed when \citet{JAZBJ18} proposed a Q-learning algorithm with \textit{upper confidence bound} (UCB) exploration for discrete tabular MDPs, and showed that it had a sample efficiency comparable to \textit{randomized least-squares value iteration} (RLSVI), a model-based method proposed by \citet{Osband16}. Later, \citet{Song19} extended that algorithm to continuous state-action spaces. Their algorithm, \textit{net-based Q-learning} (NBQL), requires discretizing the state-action space with a network of fixed step size. This creates a trade-off between memory requirements and algorithm performance. Coarser discretizations are easier to store, but the algorithm may not have sufficient resolution to achieve satisfactory performance. Finer discretizations yield better performance, but involve higher memory requirements. In order to address this trade-off, \citet{Sinclair19} proposed \textit{adaptive Q-learning} (AQL). AQL introduces adaptive discretization into NBQL, starting with a single ball covering the entire state-action space, and then adaptively discretizing it in a data-driven manner. This adaptivity ensures that the relevant parts of the state-action space are adequately partitioned, while keeping coarse discretizations in regions that are not so relevant. Shortly after, \citet{Touati20} proposed ZoomRL, a variant of AQL that keeps the regret bound, while not requiring access to a packing oracle. However, ZoomRL requires prior knowledge of the Lipschitz constant of the Q function, which may not be readily available.

The papers mentioned in the previous paragraph consider time-variant value functions and learn time-variant policies. This means that a state-action space partition needs to be kept for each (discrete) time step. However, for many practical purposes, time-invariant policies are sufficient to solve the problem satisfactorily (albeit not optimally). This is particularly true in problems with time-invariant dynamics, such as the classical \texttt{Pendulum} problem.

With this motivation, \citet{Araujo20} recently proposed \textit{single-partition adaptive Q-learning} (SPAQL), an improved version of AQL specifically tailored to learn time-invariant policies. In that paper, SPAQL is evaluated on two simple example problems as proof of concept. In this paper, SPAQL is evaluated in two classical control problems, the \pendulum~and the \cartpole, which are harder due to their more complex state and action spaces. This paper also introduces \textit{single-partition adaptive Q-learning with terminal state} (SPAQL-TS), an improved version of SPAQL, which uses concepts from control theory to achieve a better performance in the problems under study. Both SPAQL and SPAQL-TS perform better than AQL on both problems. Furthermore, SPAQL-TS manages to solve the \cartpole~problem, thus earning a place in the \gym~Leaderboard, alongside other \textit{state of the art} methods. Both SPAQL and SPAQL-TS show higher sample efficiency than \textit{trust region policy optimization} (TRPO), a standard RL algorithm known for satisfactorily solving a wide variety of control tasks \citep{duan2016benchmarking}. Furthermore, SPAQL policies are tables that map states to actions, thus being clearly interpretable. TRPO uses neural networks, which means that the learned policy is a black box. The difficulty in interpreting neural networks, as well as their large number of parameters, have stimulated the scientific community to develop more interpretable and compact policy representations (see, for example, recent work by \citet{Kubalik19}). The empirical results presented in this paper show that SPAQL and SPAQL-TS contribute to this effort of developing algorithms that efficiently learn interpretable control policies.

\subsection{Contributions}
\label{section:contributions}

This paper empirically tests AQL and SPAQL in the \pendulum~and \cartpole~control problems from the \gym~\citep{1606.01540}, classical benchmarks actively used by RL researchers \citep{doi:10.1146/annurev-control-053018-023825}. It also introduces SPAQL-TS, an improved version of SPAQL that leverages concepts from control theory to shape the value function learned by the algorithm.

To the best of the authors' knowledge, this is the first paper that empirically evaluates efficient Q-learning algorithms (\textit{i. e.}, those similar to the ones developed by \citet{JAZBJ18}, \citet{Song19}, and \citet{Sinclair19}) in classic control problems. A large number of tests with different random seeds are carried out to ensure statistical significance of the results. In both problems, SPAQL and SPAQL-TS are shown to outperform AQL, while at the same time converging faster and using a smaller number of arms (\textit{i.e.}, possible control actions).

Finally, the algorithms evaluated in this paper are compared against TRPO, an algorithm proposed by \citet{trpo}, which is known to perform well in control tasks with continuous state and action spaces \citep{duan2016benchmarking}.

\subsection{Related Work}
\label{section:related}

There exists abundant literature on the connections and applications of RL to control. For a collection of references on the topic, the reader is referred to the surveys by \citet{Kaelbling1996}, \citet{polydoros2017survey}, and \citet{BUSONIU20188}.

The papers cited by \citet{Araujo20} \citep{JAZBJ18, Song19, Sinclair19, Touati20, neustroev2020generalized, Wang2020Q-learning} either lack experimental validation or evaluate their algorithms on other types of problems (such as finding the maximum of a function, or navigating a grid world \citep{Sinclair19,neustroev2020generalized}). As stated above, this paper innovates by evaluating efficient Q-learning algorithms in control problems.

\subsection{Outline of the paper}
Section~\ref{section:preliminaries} recalls some background concepts in RL and metric spaces. Section~\ref{section:algorithm} briefly describes AQL and SPAQL. At the end of the section, SPAQL-TS, a modified version of SPAQL that incorporates concepts from control theory, is presented. The experimental setup used to evaluate the algorithm, along with the analysis of the results, can be found in Section~\ref{section:experiments}. Finally, conclusions are drawn in Section~\ref{section:conclusion}, and further work directions are proposed.

\section{Background}
\label{section:preliminaries}

For background in RL, the reader is referred to the classic book by \citet{sutton2018}. This section recalls some basic concepts of RL and metric spaces, relevant for the subsequent presentation.

\subsection{Notation used}
\label{section:preliminaries:notation}

Although they work essentialy on the same problems, the RL and control communities use different notations \citep{powell12}. This paper builds upon RL work, and the notation used in the previous papers is kept for consistency in Sections~\ref{section:preliminaries} (Background) and~\ref{section:algorithm} (Algorithm). However, since its main audience is the control community, Section
~\ref{section:experiments} (Experiments) uses the control notation. Table~\ref{table:notation} lists the symbols used to denote the problem variables in the following sections, along with their usual control counterparts.

\renewcommand{\arraystretch}{1.2}
\begin{center}
\begin{table*}[th!]
\begin{tabular}{c|c|l}
\textbf{RL} & \textbf{Control} & \textbf{Definition} \\
\textbf{Sections~\ref{section:preliminaries} and ~\ref{section:algorithm}} & \textbf{Section~\ref{section:experiments}} & \\ \hline
$x$ & $x$ & State vector of the system / Linear position (when \\
& & ambiguous, a distinction will be made) \\
 & $o$ & Observation vector of the system (simulates the sensors \\
& & available). In the RL sections, it is assumed that $o := x$ \\
$x_{ref}$ & $x_{ref}$ & Reference state being tracked in a control problem \\
$a$ & $u$ & Control action \\
$\sum_h r_h(x,a)$ & $-J$ & Cumulative reward (in control, it is more common to  \\
 &  & refer to cost $J$)  \\
$v$ & NC & Number of times a ball has been visited \\
$u$ & NC & Temperature increase factor in SPAQL \\
$d$ & NC & Temperature doubling period factor in SPAQL \\
$h$ & $t$ & Discrete time instant \\
\end{tabular}
\caption{Correspondence between the notation used in the following sections. ``NC'' (No Correspondence) means that the respective symbols only appear in Sections~\ref{section:preliminaries} and~\ref{section:algorithm} (RL).}
\label{table:notation}
\end{table*}
\end{center}

\subsection{Markov Decision Processes}

This paper adopts the Markov decision process (MDP) framework for modelling control problems. All MDPs considered have finite horizon, meaning that episodes (a simulation of the MDP for a certain number of steps) terminate after a fixed number of discrete-time steps.

Formally, an MDP is a 5-tuple $(\S, \A, H, \Pr, r)$, where:
\begin{itemize}
    \item $\S$ denotes the set of \textit{system states};
    \item $\A$ is the \textit{set of actions} of the agent interacting with the system;
     \item  $H$ is the \textit{number of steps} in each episode (also called the \textit{horizon});
     \item $\Pr$ is the \textit{transition kernel}, which assigns to each triple $(x, a, x')$ the probability of reaching state $x' \in \S$, given that action $a \in \A$ was chosen while in state $x \in \S$; this is denoted as $x' \sim \Pr(\cdot \mid x,\ a)$ (unless otherwise stated, $x'$ represents the state to which the system transitions when action $a$ is chosen while in state $x$ under transition kernel $\Pr$);
     \item $r: \S \times \A \rightarrow R \subseteq \mathbb{R}$ is the \textit{reward function}, which assigns a reward (or a cost) to each state-action pair (\citet{Sinclair19} use $R = [0, 1]$).
\end{itemize}

In this paper, the only limitation imposed on the state and action spaces is that they are bounded\footnote{It is easy to take an unbounded set and map it into a bounded one (for example, the set of real numbers can be mapped into the unit interval using the logistic function).}. Furthermore, it is assumed that the MDP has time-invariant dynamics, meaning that neither the transition kernel nor the reward function vary with time. There is previous work, such as that by \citet{NSMDP2019}, which uses the term 
``stationary'' when referring to time-invariant MDPs.

The system starts at the initial state $x_1$. At each time step $h \in \{1,..., H\}$ of the MDP, the agent receives an observation $x_{h} \in \S$, chooses an action $a_h \in \A$, receives a reward $r_h = r(x_{h},a_h)$, and transitions to state $x_{h+1} \sim \Pr(\cdot \mid x_{h},\ a_h)$. The objective of the agent is to maximize the cumulative reward $\sum_{h=1}^H r_h$ received throughout the $H$ steps of the MDP. This is achieved by learning a function $\pi : \S \rightarrow \A$ (called a \textit{policy}) that maps states to actions in a way that maximizes the accumulated rewards. This function is then used by the agent when interacting with the system. If the policy is independent of the time step, it is said to be time-invariant.

\subsection{Q-learning}

One possible approach for learning a good (eventually optimal) policy is \textit{Q-learning}. The idea is to associate with each state-action pair $(x,a)$ a number that expresses the \textbf{q}uality (hence the name Q-learning) of choosing action $a$ given that the agent is in state $x$. The value of a state $x$ at time step $h$ is defined as the expected cumulative reward that can be obtained from that state onward, under a given policy $\pi$:

\begin{equation}
V^\pi_h(x) := \Exp{\sum_{i=h}^H r(x_{i}, \pi(x_{i})) ~\Big|~ x_h = x}.
\end{equation}

This is equivalent to averaging all cumulative rewards that can be obtained under policy $\pi$ until the end of the MDP, given that at time step $h$ the system is in state $x$. By taking into account all rewards until the end of the MDP, the value function provides the agent with information regarding the rewards in the long run. This is relevant, since the state-action pairs yielding the highest rewards in the short run may not be those that yield the highest cumulative reward. However, if $H$ is large enough and there is a large number of state-action pairs, it may be impossible (or impractical) to compute the actual value of $V^\pi_h(x)$ for all $1 \leq h \leq H$ and $x \in \S$, and some approximation has to be used instead. For example, a possible estimator samples several paths at random, and averages the cumulative rewards obtained \citep{sutton2018}.

The \textit{value function} $V^\pi_h : \S \rightarrow \mathbb{R}$ provides information regarding which states are more desirable. However, it does not take into account the actions that the agent might choose in a given state. The \textit{Q function},

\begin{equation}
Q^\pi_h(x, a) := r(x,a) + \Exp{\sum_{i=h+1}^H r(x_{i}, \pi(x_{i})) ~\Big|~ x_h = x,\ a_h = a} = r(x,a) + \Exp{V^\pi_h(x')~\Big|~ x,\ a},
\end{equation}
takes this information into account since its domain is the set of all possible state and action pairs. This allows the agent to rank all possible actions at state $x$ according to the corresponding values of Q, and then make a decision regarding which action to choose. Notice that the second expectation in the previous equation is with respect to $x'  \sim \Pr(\cdot \mid x,\ a)$.

The functions $V^\star_h(x) = \sup_\pi V^\pi_h(x)$ and $Q^{\star}_h(x, a) = \sup_\pi Q^\pi_h(x, a)$ are called the \textit{optimal value function} and the \textit{optimal Q function}, respectively. The policies $\pi^\star_V$ and $\pi^\star_Q$ associated with $V^\star_h(x)$ and $Q^{\star}_h(x, a)$ are one and the same, $\pi^\star_V=\pi^\star_Q=\pi^\star$; this policy is called the \textit{optimal policy} \citep{sutton2018}. There may be more than one optimal policy, but they will always share the same optimal value and optimal Q function. These functions satisfy the so-called \textit{Bellman equation}

\begin{equation}
\label{eq:bellman}
    Q^{\star}_h(x, a) = r(x, a) + \mathbb{E}\left[V^{\star}_h(x') \mid x, a\right].%,
\end{equation}

Q-learning is an algorithm for computing estimates $\Qhat{h}{}$ and $\Vhat{h}{}$ of the Q function and the value function, respectively. The updates to the estimates at each time step are based on Equation~\ref{eq:bellman}, according to

\begin{equation}
\Qhat{h}{}(x, a) \leftarrow (1-\alpha)\Qhat{h}{}(x, a) + \alpha\left(r(x,a) + \Vhat{h}{}(x')\right),
\end{equation}
where $\alpha \in [0, 1]$ is the  \textit{learning rate}. This update rule reaches a fixed point (stops updating) when the Bellman equation is satisfied, meaning that the optimal functions were found. Actions are chosen greedily according to the $\argmax$ policy

\begin{equation}
\pi(x) = \argmax_a \Qhat{h}{}(x, a).
\end{equation}

The greediness of the $\argmax$ policy may lead the agent to become trapped in local optima. To escape from these local optima, stochastic policies such as $\varepsilon$-greedy or Boltzmann exploration can be used. The $\varepsilon$-greedy policy chooses the greedy action with probability $1-\varepsilon$ (where $\varepsilon$ is a small positive number), and picks any action uniformly at random with probability $\varepsilon$ \citep{sutton2018}. Boltzmann exploration transforms the $\Qhat{h}{}$ estimates into a probability distribution (using softmax), and then draws an action at random. It is parametrized by a temperature parameter $\tau$, such that, when $\tau \rightarrow 0$, the policy tends to $\argmax$, whereas, for $\tau \rightarrow +\infty$, the policy tends to one that picks an action uniformly at random from the set of all possible actions.

Finding and escaping from local optima is part of the exploration/exploitation trade-off. Another way to deal with this trade-off is to use \textit{upper confidence bounds} (UCB). Algorithms that use UCB add an extra term $b(x,a)$ to the update rule

\begin{equation}
\Qhat{h}{}(x, a) \leftarrow (1-\alpha)\Qhat{h}{}(x, a) + \alpha\left(r(x,a) + \Vhat{h}{}(x') + b(x,a)\right),
\end{equation}
which models the uncertainty of the Q function estimate. An intuitive way of explaining UCB is to say that the goal is to choose a value of $b(x,a)$ such that, with high probability, the actual value of $\Qhat{h}{}(x, a)$ lies within the interval $[\Qhat{h}{}(x, a), \Qhat{h}{}(x, a)+b(x,a)]$. A very simple example is $b(x,a) = 1/n_{(x,a)}$, where $n_{(x,a)}$ is the number of times action $a$ was chosen while the agent was in state $x$. As training progresses and $n_{(x,a)}$ increases, the uncertainty associated with the estimate of $\Qhat{h}{}(x, a)$ decreases. This decrease affects the following training iterations since, eventually, actions that have been less explored will have higher $\Qhat{h}{}(x, a)$ due to the influence of term $b(x,a)$, and will be chosen by the greedy policy. This results in exploration. 

In the finite-horizon case (finite $H$), the optimal policy usually depends on the time step \citep{busoniu2010reinforcement}. One of the causes for this dependency is the constraint $V_{H+1}^{\pi}(x)=0$ for all $x \in \mathcal{S}$ (as mentioned by \citet{Sinclair19} and \citet[Section 6.5]{sutton2018}), since it forces the value function to be time-variant. The SPAQL algorithm, which learns time-invariant policies, deals with this by ignoring this constraint. For more information, refer to Section~\ref{sec:alg:spaql}.

\subsection{Metric spaces}

A metric space is a pair $(X, \D)$ where $X$ is a set and $\D: X \times X \rightarrow \mathbb{R}$ is a function (called the \textit{distance function}) satisfying the following properties:

\begin{equation}
\begin{array}{lll}
    \D(x,x) = 0&&\forall x \in X\\
    \D(x,y) = \D(y,x)&&\forall x,\ y \in X\\
    \D(x,z) \leq \D(x,y) + \D(y,z)&&\forall x,\ y,\ z \in X\\
    \D(x,y) \geq 0&&\forall x,\ y \in X.
\end{array}
\end{equation}

A ball $B$ with center $x$ and radius $r$ is the set of all points in $X$ which are at a distance strictly lower than $r$ from $x$, $B(x,r) = \{b \in X : \D(x, b) < r\}$. The diameter of a ball is defined as $\text{diam}(B) = \sup_{x, y \in B} \D(x, y)$. The diameter of the entire space is denoted $d_{max} = \text{diam}(X)$. Next, the concepts of \textit{covering}, \textit{packing}, and \textit{net} are recalled.

\begin{definition}[\citet{Sinclair19}]
An \textit{$r$-covering} of $X$ is a collection of subsets of $X$ that covers $X$ (\textit{i.e.}, any element of $X$ belongs to the union of the collection of subsets) and such that each subset has diameter strictly less than $r$.
\end{definition}

\begin{definition}[\citet{Sinclair19}]
\label{definition:net}
{A set of points $\P \subset X$} is an \textit{$r$-packing} if the distance between any two points in $\P$ is at least $r$.  An $r$-\textit{net} of $X$ is an $r$-packing such that $X \subseteq \cup_{x \in \P} B(x, r)$.
\end{definition}

\section{Algorithms}
\label{section:algorithm}

\begin{algorithm*}[t!]
	\caption{Adaptive $Q$-learning (\cite{Sinclair19})}
	\label{aql_alg}
	\begin{algorithmic}[1]
		\Procedure{Adaptive $Q$-learning}{$\S, \A, \D, H, K, \pfail$}
			\State Initiate $H$ partitions $\P_h^1$ for $h = 1, \ldots, H$ each containing a single ball with radius $d_{max}$ and $\Qhat{h}{1}$ estimate $H$
			\For{each episode $k \gets 1, \ldots K$}
				\State Receive initial state $x_1^k$
				\For{each step $h \gets 1, \ldots, H$}
					\State Select the ball $B_{sel}$ by the selection rule $B_{sel} = \displaystyle \argmax_{B \in \texttt{RELEVANT}_h^k(x_h^k)} \Qhat{h}{k}(B)$
					\State Select action $a_h^k = a$ for some $(x_h^k, a) \in \dom{B_{sel}}$
					\State Play action $a_h^k$, receive reward $r_h^k$ and transition to new state $x_{h+1}^k$
					\State Update Parameters: $v = n_h^{k+1}(B_{sel}) \gets n_h^{k}(B_{sel}) + 1$ \\ 
					\State $\Qhat{h}{k+1}(B_{sel}) \gets (1 - \alpha_v)\Qhat{h}{k}(B_{sel}) + \alpha_v(r_h^k + \Vhat{h+1}{k}(x_{h+1}^k) + b_{\xi}(v))$ where \\
					\State $\Vhat{h+1}{k}(x_{h+1}^k) = \min(H, \displaystyle \max_{B \in \texttt{RELEVANT}_{h+1}^k(x_{h+1}^k)} \Qhat{h+1}{k}(B))$ (see Section~\ref{sec:alg:aql}, Equation~\ref{eqn:future_value})
					\If{$n_h^{k+1}(B_{sel}) \geq \left( \frac{d_{max}}{r(B_{sel})}\right)^2$}
					  \textproc{Split Ball}$(B_{sel}, h, k)$
					\EndIf
				\EndFor
			\EndFor
		\EndProcedure
		\Procedure{Split Ball}{$B$, $h$, $k$}
		    \State Set $B_1, \ldots B_n$ to be an $\frac{1}{2}r(B)$-packing of $\dom{B}$, and add each ball to the partition $\P_h^{k+1}$ (see Definition~\ref{definition:net})
		    \State Initialize parameters $\Qhat{h}{k+1}(B_i)$ and $n_h^{k+1}(B_i)$ for each new ball $B_i$ to inherit values from the parent ball $B$
		 \EndProcedure
	\end{algorithmic}
\end{algorithm*}

For a detailed description of both AQL and SPAQL, please refer to the original papers by \citet{Sinclair19} and \citet{Araujo20}. This section offers a brief overview of both algorithms, but its objective is not to be exhaustive.

\subsection{Adaptive Q-learning}
\label{sec:alg:aql}
Adaptive Q-learning (AQL) is a Q-learning algorithm for finite-horizon MDPs with continuous state-action spaces. It keeps a state-action space partition for each time step. While this is fundamental when dealing with time-variant MDPs, for time-invariant ones it may result in an excessive use of memory resources.

The pseudocode for AQL is shown in Algorithm~\ref{aql_alg}. For each variable, superscripts denote the current episode (training iteration) $k$, and subscripts denote the time step $h$. Intuitively, the algorithm partitions a collection $\P^1 = \{\P^1_h: h=1,\ldots,H \}$ of initial balls (one per time step $h$), each containing the entire state-action space, into smaller balls. For each ball $B$, it keeps an estimate of the value of the Q-function, denoted by $\Qhat{h}{k}(B)$ (estimate of $\Qhat{}{}$ at time step $h$ of episode $k$).

A ball $B_i$ is said to be relevant for the current state $x_h^k$ (state at time step $h$ of episode $k$) if there exists at least one pair $(x_h^k, a) \in \S \times A$ such that $(x_h^k, a) \in \dom{B_i}$, where $\dom{B_i}$ is the domain of ball $B_i$. The domain of a ball is defined as

\begin{equation}\label{eqn:domain}
\dom{B} = B \setminus \left( \bigcup_{B' \in \P : r(B') < r(B)} B'\right).
\end{equation}
In other words, the domain of a ball $B$ is the set of all points $b \in B$ which are not contained inside any other ball of strictly smaller radius than $r(B)$. This concept is illustrated in Appendix~\ref{sec:domain}. In the examples presented in this thesis, it is ensured that either $\dom{B} = B$, or $\dom{B} = \varnothing$. Each case corresponds to before and after splitting ball $B$, respectively.

The set of all relevant balls at a given time step $h$ of a given episode $k$ is denoted by $\texttt{RELEVANT}_h^k(x_h^k)$ (or simply $\texttt{RELEVANT}(x_h^k)$). Given the current state $x_h^k$, the algorithm selects the relevant ball with the highest estimate of the Q-function, denoted by $B_{sel}$. It then picks an action uniformly at random from within that ball. The number of times a ball has been visited at time step $h$ is denoted $n_h^k$. This number carries across training iterations, as per line 9 of Algorithm~\ref{aql_alg}. If the number of times a ball has been visited is above a certain threshold of times, it is then partitioned into smaller balls, thus refining the state-action space partition in the regions where most visits have occurred. This threshold is defined as $ \left(d_{max}/r(B_{sel})\right)^2$, where $d_{max}$ is the radius of the initial ball covering the entire state-action space, and $r(B)$ is the radius of the currently selected ball. If each child ball has half the radius of its parent, then $r(B_{sel}) = d_{max}/2^n$, where $n$ is the number of times that ball $B$'s parents have been split. Substituting in the threshold formula, the equivalent expression $4^n$ is obtained \citep{Sinclair19}.

The update rule for the estimates of $\Qhat{h}{k}$ is given by

\begin{equation}\label{eqn:qval_update}
\Qhat{h}{k+1}(B_{sel}) \gets (1 - \alpha_v)\Qhat{h}{k}(B_{sel}) + \alpha_v(r_h^k + \Vhat{h+1}{k}(x_{h+1}^k) + b_{\xi}(v)),
\end{equation}
where $b_{\xi}(v)$ is the UCB bonus term, $v$ is the number of times that the currently selected ball $B_{sel}$ has been \textbf{v}isited, and $\alpha_v$ is the learning rate. Both $b_{\xi}(v)$ and $\alpha_v$ are defined below. The $k+1$ superscript in $\Qhat{h}{k+1}$ means that the update to the estimate carries to the next episode, not having any effect on the current one (similar to what was previously explained for $n_h^k$). 

The value function is defined as

\begin{equation}\label{eqn:future_value}
\Vhat{h+1}{k}(x_{h+1}^k) = \min(H, \displaystyle \dsmax_{B \in  \texttt{RELEVANT}(x_{h+1}^k)} \Qhat{h+1}{k}(B))
\end{equation}
for all non-terminal states. The value of the terminal state, $\Vhat{H+1}{k}(x)$ (where $H$ is the horizon), is defined to be $0$ for all states $x \in \S$.

The trade-off between exploration and exploitation is handled using an upper confidence bound (UCB) of the estimate of the Q-function, $b_{\xi}(v)$, defined as

\begin{equation}\label{eqn:conf_radius_practical}
b_{\xi}(v) = \frac{\xi}{\sqrt{v}},
\end{equation}
where $\xi$ is a user-defined parameter. \citet{Sinclair19} provide an expression for the value of $\xi$,

\begin{equation}\label{eqn:conf_radius}
\xi = 2 \sqrt{H^3 \log\left(\frac{4HK}{\pfail}\right)} + 4Ld_{max},
\end{equation}
where $K$ is the number of training iterations, $\pfail$ is a parameter related with the high-probability regret bound \citep{Sinclair19}, $L$ is the Lipschitz constant of the Q-function, and $d_{max}$ is the radius of the initial ball covering the entire state-action space. However, as pointed out by \citet{Araujo20}, setting $\xi$ according to Equation~\ref{eqn:conf_radius} may not yield the best results in practice. In their implementation, \citet{Sinclair19} tune and set the value of $\xi$ by hand (Equation~\ref{eqn:conf_radius} is used to calculate the regret bounds of AQL).

The learning rate $\alpha_v$ is set according to

\begin{equation}\label{eqn:learning_rate}
\alpha_v = \frac{H+1}{H+v}.
\end{equation}

The previous description of the algorithm can be summed up as the application of three rules:

\begin{itemize}
	\item \textbf{Selection rule}: given the current state $x_h^k$, select the ball with the highest value of $\Qhat{h}{k}$ (out of the ones which are relevant), and from within that ball pick an action $a_h^k$ uniformly at random.
	\item \textbf{Update parameters}: increment $n_h(B)$, the number of times ball $B$ has been visited (the training iteration $k$ is not relevant and can be dropped), and update $\Qhat{h}{k+1}(B)$ according to Equation~\ref{eqn:qval_update} ($k+1$ is notation for carrying the Q-function estimate over to the next episode).
	\item \textbf{Re-partition the space}: when a ball $B$ has been visited more than $ \left(d_{max}/r(B)\right)^2$ times, cover it with a $\frac{1}{2}r(B)$-Net of $B$; each new ball inherits the number of visits and the Q-function estimate from its parent ball.
\end{itemize}

\subsection{Single-partition adaptive Q-learning}
\label{sec:alg:spaql}

\begin{algorithm}[t!]
    \caption{Auxiliary functions for SPAQL \citep{Araujo20}}
    \label{alg:aux}
    \begin{algorithmic} [1]
        \Procedure{Evaluate Agent}{$\P, \S, \A, H, N$}
	    \State Collect $N$ cumulative rewards from \textproc{Rollout}$(\P, \S, \A, H)$
	    \State Return the average cumulative reward
		\EndProcedure
		\Procedure{Rollout}{$\P, \S, \A, H$}
		\State Receive initial state $x_1$
		\For{each step $h \gets 1, \ldots, H$}
		    \State Pick the ball $B_{sel} \in \P$ by the selection rule $B_{sel} = \displaystyle \argmax_{B \in \texttt{RELEVANT}(x_h)} \Qhat{}{}(B)$
			\State Select action $a_h = a$ for some $(x_h, a) \in \dom{B_{sel}}$
			\State Play action $a_h$, record reward $r_h$ and transition to new state $x_{h+1}$
		\EndFor
		\State Return cumulative reward $\displaystyle \sum_h r_h$
		\EndProcedure
		\Procedure{Split Ball}{$B$}
		    \State Set $B_1, \ldots B_n$ to be an $\frac{1}{2}r(B)$-packing of $\dom{B}$, and add each ball to the partition $\P$ (see Definition~\ref{definition:net})
		    \State Initialize parameters $\Qhat{}{}(B_i)$ and $n(B_i)$ for each new ball $B_i$ to inherit values from the parent ball $B$
		 \EndProcedure
		 \Procedure{Boltzmann Sample}{$\Bhat$, $\tau$}
		    \State Normalize values in $\Bhat$ by dividing by $\max(\Bhat)$ (this helps to prevent overflows)
		    \State Sample a ball $B$ from $\Bhat$ by drawing a ball at random with probabilities following the distribution
		    \[
		    P(B = B_i) \sim \exp\left(\Qhat{}{}(B_{i})/\tau\right)
		    \]
		 \EndProcedure
	\end{algorithmic}
\end{algorithm}

\begin{algorithm*}[t!]
	\caption{Single-partition adaptive $Q$-learning \citep{Araujo20}}
	\label{spaql_alg}
	\begin{algorithmic}[1]
		\Procedure{Single-partition adaptive $Q$-learning}{$\S, \A, \D, H, K, N, \xi, \taumin, u, d$}
			\State Initialize partitions $\P$ and $\P'$ containing a single ball with radius $d_{max}$ and $\Qhat{}{} = H$
			\State Initialize $\tau$ to $\taumin$
			\State Calculate agent performance using \textproc{Evaluate Agent}($\P, \S, \A, H, N$) (Algorithm~\ref{alg:aux})
			\For{each episode $k = 1, \ldots, K$}
				\State Receive initial state $x_1^k$
				\For{each step $h = 1, \ldots, H$}
				    \State Get a list $\Bhat$ with all the balls $B \in \P'$ which contain $x_h^k$
					\State Sample the ball $B_{sel}$ using \textproc{Boltzmann Sample}$(\Bhat, \tau)$ (Algorithm~\ref{alg:aux})
					\State Select action $a_h^k = a$ for some $(x_h^k, a) \in \dom{B_{sel}}$
					\State Play action $a_h^k$, receive reward $r_h^k$ and transition to new state $x_{h+1}^k$
					\State Update Parameters: $v = n(B_{sel}) \gets n(B_{sel}) + 1$ \\ 
					\State $\Qhat{}{}(B_{sel}) \gets (1 - \alpha_v)\Qhat{}{}(B_{sel}) + \alpha_v(r_h^k + \Vhat{}{}(x_{h+1}^k) + b_{\xi}(v))$ where \\
					\State \indent $\Vhat{}{}(x_{h+1}^k) = \min(H, \displaystyle \dsmax_{B \in  \texttt{RELEVANT}(x_{h+1}^k)} \Qhat{}{}(B))$ (see Section~\ref{sec:alg:spaql})
					\If{$n(B_{sel}) \geq \left( \frac{d_{max}}{r(B_{sel})}\right)^2$}
					  \textproc{Split Ball}$(B_{sel})$ (Algorithm~\ref{alg:aux})
					\EndIf
				\EndFor
				\State Evaluate the agent using \textproc{Evaluate Agent}($\P', \S, \A, H, N$) (Algorithm~\ref{alg:aux})
				\If{agent performance improved}
				    \State Copy $\P'$ to $\P$ (keep the best agent)
				    \State Reset $\tau$ to $\taumin$
				    \State Decrease $u$ using some function of $d$ (for example, $u \gets u^d$, assuming $d < 1$)
				\Else
				    \State Increase $\tau$ using some function of $u$ (for example, $\tau \gets u\tau$)
				    \If{more than two splits occurred}
				        \State Copy $\P$ to $\P'$ (reset the agent)
				        \State Reset $\tau$ to $\taumin$
				    \EndIf
				\EndIf
			\EndFor
		\EndProcedure
    \end{algorithmic}
\end{algorithm*}

Before describing SPAQL, it is necessary to define several auxiliary functions, the pseudocode for which is shown in Algorithm~\ref{alg:aux}. Function \textproc{Evaluate Agent} is used to evaluate the performance of an agent; it does so by running the agent on the episode (\textproc{Rollout}) $N$ times, and then averaging the $N$ cumulative rewards. Function \textproc{Split Ball} is similar to the one used in AQL (Algorithm~\ref{aql_alg}), the only difference being that the dependence on the time step $h$ is dropped. Finally, procedure \textproc{Boltzmann Sample} is used to do Boltzmann exploration, as described below.

SPAQL (Algorithm~\ref{spaql_alg}) is an improved version of AQL specifically tailored to learn time-invariant policies. This means that only one estimate of the value and Q-function is kept, instead of one per time step ($\Qhat{h}{k} := \Qhat{}{k}$ and $\Vhat{h}{k} := \Vhat{}{k}$). Besides this, the main differences to AQL are the definition of the value function, and the way that exploration is balanced with exploitation.

The boundary condition imposed on the value of the terminal state, $\Vhat{H+1}{k}(x) = 0$, for any state $x$, forces the value function to be time-variant. SPAQL circumvents this by eliminating this condition and using the same definition of value function as if the state was a non-terminal one.

Exploration is performed using a mix of upper confidence bounds UCB and Boltzmann exploration with a cyclical adaptive temperature $\tau$ schedule. The bonus term of the UCB, $b_{\xi}(v)$, is defined according to Equation~\ref{eqn:conf_radius_practical}, with parameter $\xi$ set by the user.

In the first training iteration, the temperature $\tau$ is at its lowest value ($\taumin$), making the policy to be $\argmax$ (as in AQL). At the end of each training iteration, the policy is evaluated. The best performing policy found so far is kept in storage, and training occurs on a copy of it. Every time a better policy is found, it overwrites the previous best one. For each training iteration on which performance does not improve, the temperature parameter is increased by a factor $u$. If the performance does not improve for an (implicitly) user-defined number of iterations, the policy becomes random. Each trainee policy is allowed to split existing balls twice. If performance did not improve after the second split, the trainee policy is reset to the best one found so far, and the temperature parameter $\tau$ is also reset to its lowest value, $\taumin$. Every time a ball is split, more visits are required to split its children, and thus the parameter $u$ is decreased by a factor $d$ every time the agent improves. This way, more iterations are required in order for the policy to become random.

Summarizing the ways in which SPAQL differs from AQL, it follows the three rules described in the previous section, plus two new rules related with exploration:

\begin{itemize}
	\item \textbf{Adapt the temperature}: increase the temperature at every training iteration in which no improvement was detected, and reset it to $\taumin$ when performance improves.
	\item \textbf{Reset the agent}: if performance has not improved after two balls were split, reset the agent to the best one found so far and continue training.
\end{itemize}

\subsection{Single-partition adaptive Q-learning with terminal state}

\label{sec:spaql-ts}

When designing a controller, the goal is to have the final state of the system to be as close as possible to a user-defined state (reference). Denoting this reference state by $x_{ref}$, it is possible to modify the value function used in SPAQL to take this information into account. The intuition behind this idea is to decrease the value of the states based on their distance to the reference state. There are several possible ways to do this, depending on the properties sought for the new value function. One such possibility is to weight the value function with a Gaussian with mean $x_{ref}$ and user-defined standard deviation $\lambda$. The resulting value function,

\begin{equation}\label{eqn:future_value_control}
\Vhat{}{k}(x_{h+1}^k) = \exp \left(-\left(\frac{\D(x_h^k, x_{ref})}{\lambda}\right)^2\right) \min\left(H, \displaystyle \dsmax_{B \in  \texttt{RELEVANT}(x_{h+1}^k)} \Qhat{}{}(B)\right),
\end{equation}
weights the SPAQL value function according to the distance $\D(x_h^k, x_{ref})$ to the reference state (error). The weight given to the error is controlled by the parameter $\lambda > 0$. Setting $\lambda$ to a low value forces the algorithm to search for policies that minimize the error as much as possible, while setting $\lambda$ to a large value allows the algorithm to search for more tolerant policies.

This modified version of SPAQL is referred to as SPAQL with terminal state (SPAQL-TS).

This very simple idea allows the introduction of some domain knowledge into the concept of quality. Instead of defining quality solely as a blind search for high cumulative rewards, this modified value function introduces the notion of error, and correlates high-quality states with low error values. In principle, this should allow for more efficient training.

\section{Experiments}
\label{section:experiments}

In this section, AQL, SPAQL, and SPAQL-TS are tested and compared on the \texttt{Pendulum} and \texttt{CartPole} problems of \texttt{OpenAI Gym} \citep{1606.01540}. To close the section, these three algorithms are compared against TRPO \citep{trpo}, an RL algorithm known to perform well on these two problems \citep{duan2016benchmarking}.

Throughout this section, the control notation is adopted. Refer to Table~\ref{table:notation} in Section~\ref{section:preliminaries:notation} for the correspondence between the notation used in the previous sections and the notation used in the ones that follow. \blfootnote{The code associated with all experiments presented in this paper is available at \url{https://github.com/jaraujo98/SinglePartitionAdaptiveQLearning}}

\subsection{Implementation}
\label{sec:implementation}

The state-action spaces for the \pendulum~and \cartpole~problems are described in Sections~\ref{section:experiments_set_up_pendulum} and~\ref{section:experiments_set_up_cartpole}, respectively. The data structures implemented to store the partitions used by the agents assume that the state-action pairs would be converted into a standard space (namely $[-1,1]^3\times[-1,1]$ for the \pendulum, and $[-1,1]^4\times\{0,1\}$ for the \cartpole). The reason for using these standard spaces is that both the state-action space of the \pendulum~and the state space of the \cartpole~problems can be tightly covered by a ball with radius $1$ centered at the origin. The finite action set of the \cartpole~problem, $\{0,1\}$, has to be dealt with separately. After checking if the state is within the selected ball, the implementation asserts that the action chosen is contained within the action set associated with the same ball. If the ball only has one possible action (the action set is a singleton), the state-action pair can only be contained within that ball if the action is the one associated with that ball. As \citet{Sinclair19} and \citet{Araujo20} propose, a tree data structure is used to store the partitions, and ball selection is done using a recursive algorithm.

\subsection{Procedure and parameters}

\label{sec:exp_proc_param}

Following \citet{Araujo20}, the effect of the scaling parameter $\xi$ on AQL and SPAQL is studied. For each algorithm and scaling value, $20$ agents were trained for $100$ iterations. By definition, the episode length on both systems is $200$ \citep{1606.01540}. The number of rollouts $N$ used to evaluate each agent was set to $20$. The scaling values used are based on the ones listed by \cite{Araujo20}, scaled according to the change in horizon length (from $H=5$ in \citep{Sinclair19} to $H=200$ in this paper),

\begin{equation}
\xi \in \{0.4, 4, 10, 20, 30, 40, 50, 60, 70, 80, 120, 160\}.
\end{equation}
As a baseline, the effect of setting the scaling parameter $\xi$ to $0$ is also studied. This disables the upper confidence bounds, turning both algorithms into AQL algorithms with Boltzmann exploration.

For SPAQL and SPAQL-TS, the values of $\taumin$, $u$ (SPAQL parameter, not to be confused with the control action), and $d$ are set to $0.01$, $2$, and $0.8$, respectively. For SPAQL-TS, the value of $\lambda$ is set to $1.2$. Table \ref{tab:scaling_exp_parameters} contains the parameters used in the experiments reported in Figures~\ref{fig:pendulum_scaling_results} and~\ref{fig:cartpole_scaling_results}. The average cumulative reward at the end of training was calculated over the several agents, along with the $95\%$ confidence intervals.

After the values of $\xi$ which resulted on higher average cumulative rewards were identified, a respective number of $20$ agents were trained for $2000$ iterations. In order to check for problem resolution (refer to Section~\ref{section:experiments_set_up_cartpole}), the \cartpole~agents were evaluated $100$ times instead of $20$. A summary of these parameters is listed in Table~\ref{tab:exp_parameters}.

\begin{table}[ht]
\centering
\begin{tabular}{l|c}
\hline
\multicolumn{1}{c}{Parameter} & Value \\\hline
Episode length $H$               & 200     \\
Number of scaling episodes (training iterations) $K$           & 100  \\
Number of agents              & 20   \\
Number of evaluation rollouts $N$             & 20   \\\hline
\end{tabular}
\caption{Parameters for the scaling experiments}
\label{tab:scaling_exp_parameters}
\end{table}

\begin{table}[ht]
\centering
\begin{tabular}{l|c}
\hline
\multicolumn{1}{c}{Parameter} & Value \\\hline
Episode length $H$               & 200     \\
Number of episodes (training iterations) $K$           & 2000  \\
Number of agents              & 20   \\
Number of evaluation rollouts $N$ (\pendulum)            & 20   \\
Number of evaluation rollouts $N$ (\cartpole)            & 100   \\\hline
\end{tabular}
\caption{Parameters for the experiments after choosing $\xi$}
\label{tab:exp_parameters}
\end{table}

\subsection{Pendulum}

\subsubsection{Setup}
\label{section:experiments_set_up_pendulum}

The objective is to drive a pendulum with length $l=1$m and mass $m=1$kg to the upright position by applying a torque on a joint at one of its ends. The state of the system is the angular position of the pendulum with the vertical axis, $\theta$, and the angular velocity $\dot{\theta}$ of the pendulum. This can be written compactly as $x = [\theta, \dot{\theta}]^T$. The control action $u$ is the value of the torque applied (which can be positive or negative, to indicate direction). The observation $o = [c_{\theta}, s_{\theta}, \dot{\theta}]^T$ is a $3D$ vector whose entries are the cosine ($c_{\theta}$) and sine ($s_{\theta}$) of the pendulum's angle, and the angular velocity \citep{1606.01540}. This vector simulates the sensors available to measure the state variables of the system, and is part of the definition of the \gym~problem. There is no distinction between state and observation spaces in Section~\ref{section:algorithm}, but it is clear that \citet{Sinclair19} and \citet{Araujo20} have the observation space in mind when they mention state space (since the samples used by the algorithms must necessarily come from the observation space). The angular velocity saturates at $-8$\ rad/s and $8$\ rad/s, and the control action saturates at $-2$\ Nm and $2$\ Nm. The cost to be minimized (which, from an RL point of view, is equivalent to a negative reward) is given by \citep{1606.01540}

\begin{equation}
J = \sum_{t=1}^H\text{normalize}(\theta_t)^2+ 0.1\dot{\theta}_t^2 + 0.001(u_t^2),
\end{equation}
where normalize($\cdot$) is a function that maps an angle into its principal argument (\textit{i. e.}, a value in $]-\pi, \pi]$).

AQL and SPAQL were designed with the assumption that the rewards were in the interval $[0, 1]$ (which implies that $\Vhat{}{k}(o) \in [0, H]$, for all observations $o$). The effect of different cost structures is assessed by training the algorithms on the actual system (with rewards ranging from approximately $-16$ to $0$) and on a system with rewards scaled to be in the interval $[0, 1]$. The UCB bonus term (Equation~\ref{eqn:conf_radius_practical}) can be expected to play an important part in the original problem, since it may balance the negative reward by making the term $r^k_t + b_{\xi}(v)$ in Equation~\ref{eqn:qval_update} always positive during training. 

The state-action space of this problem (as observed by the agent) is

\[
\S \times \A = \left([-1, 1] \times [-1, 1] \times [-8, 8]\right) \times [-2, 2].
\]

In order to simplify the implementation (Section~\ref{sec:implementation}), this space is mapped to a standard space ($[-1,1]^4$). The angular velocity ($\dot{\theta} \in [-8, 8]$) is divided by $8$, and then passed to the agent. The agent then picks a control action $u \in [-1, 1]$. Before this action is sent to the simulation, it is multiplied by $2$.

The $\infty$ product metric (used by the agent) is written as

\begin{equation}
\label{metric_pendulum}
    \D((x,u), (x',u')) = \max\bigl\{|c_{\theta} - c_{\theta'}|, |s_{\theta} - s_{\theta'}|, \frac{|\dot{\theta} - \dot{\theta'}|}{8}, |u-u'|\bigr\}.
\end{equation}
Since the agent only sees the state-action pairs in the standard space $[-1, 1]^4$, the control action it picks is already contained in the interval $[-1, 1]$. It is the interface's task to scale this control action back to the action space of the problem. Since the control action is already correctly scaled, the component associated with the control action $u$ in Equation~\ref{metric_pendulum} does not have to be divided by $2$.

For SPAQL-TS, the reference observation $o_{ref}$ (denoted $x_{ref}$ in Section~\ref{sec:spaql-ts}) is set to $[1, 0, 0]^T$, corresponding to reference state $[\theta_{ref}, \dot{\theta}_{ref}]^T = [0, 0]^T$.

\subsubsection{Results}
\label{pendulum_experimental_results}

\begin{figure*}[t!]
\centering
\includegraphics[width=0.75\columnwidth]{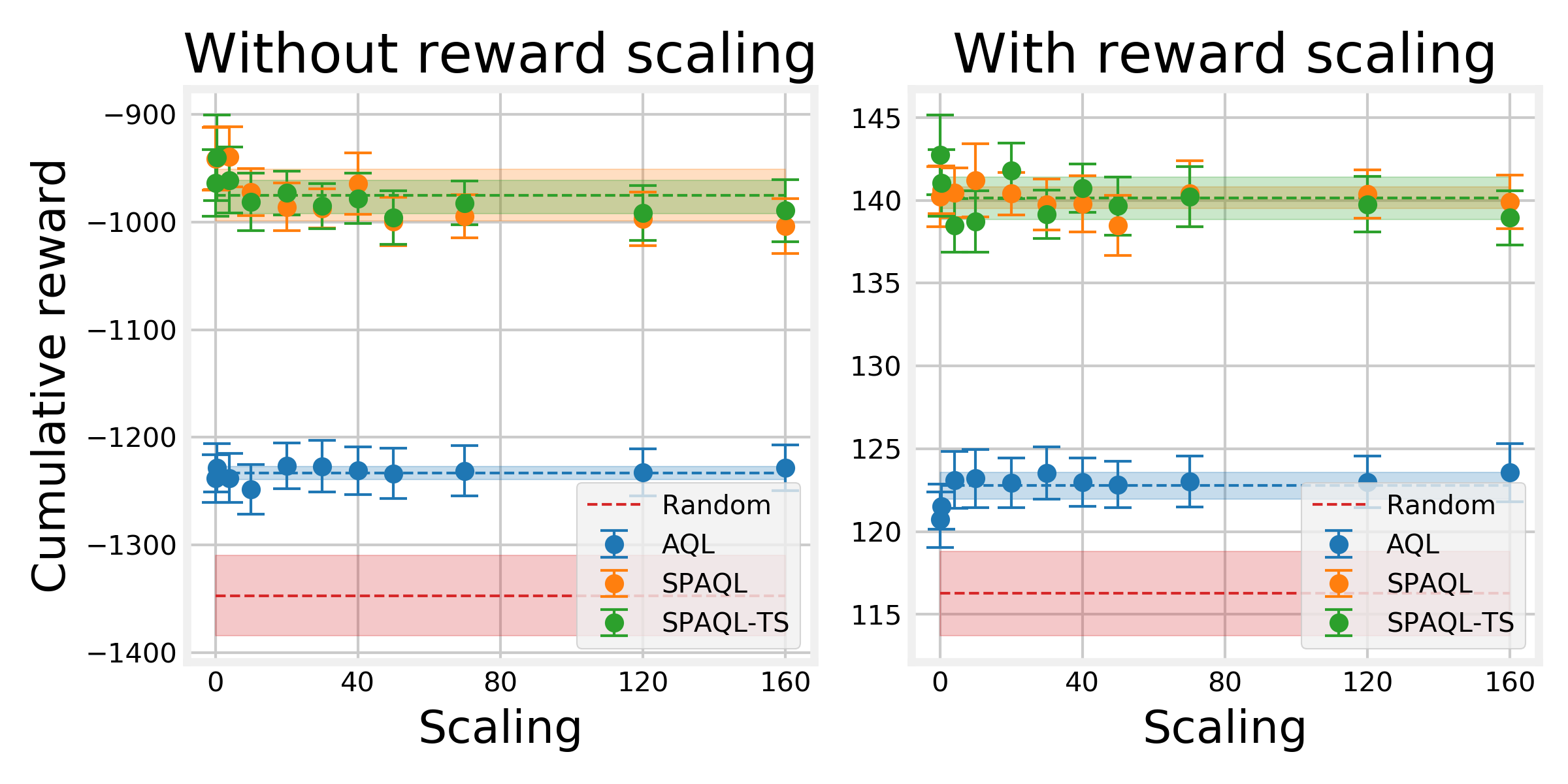}
\caption{
Comparison of different scaling parameter values on the average cumulative reward for the \pendulum~system (with and without reward scaling). Each dot corresponds to the average of the rewards obtained by the $20$ agents after $100$ training iterations, and the error bars display the corresponding $95\%$ confidence interval. Dashed lines represent the average cumulative reward calculated over the scaling values listed in Section~\ref{sec:exp_proc_param}, and the shaded areas represent the corresponding standard deviation.
}
\label{fig:pendulum_scaling_results}
\end{figure*}

The results in Figure~\ref{fig:pendulum_scaling_results} show that the scaling parameter has little impact on the three algorithms. Setting the scaling values $\xi$ to $0$ seems to be beneficial to both SPAQL and SPAQL-TS, meaning that Boltzmann exploration suffices to handle this problem. Although $100$ iterations is a small number, it is already clear that both algorithms are learning (\textit{i. e.}, the policies perform better than random ones), with SPAQL learning more than AQL. There is no difference between the two variants of SPAQL. Reward scaling does not greatly impact the results. Despite the difference in cumulative reward axes scales, the relative positions of the different algorithms in the plots with and without reward scaling remain the same. A possible explanation for the similarity between the learning curves with and without reward scaling is that the optimistic initialization of the Q-function manages to absorb the effect of the negative rewards.

\begin{figure*}[t!]
\centering
\includegraphics[width=0.75\columnwidth]{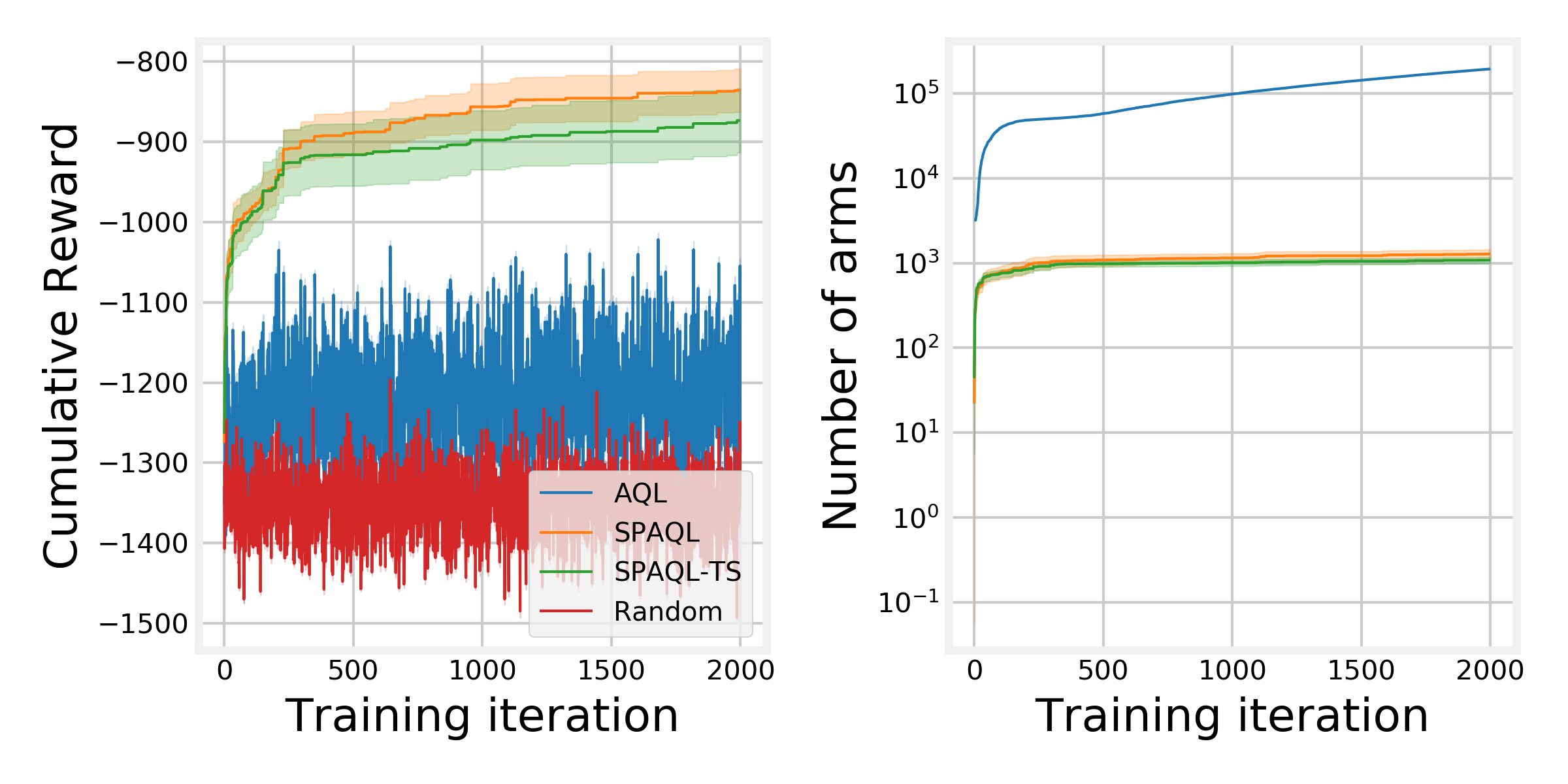}
\caption{Average cumulative rewards and number of arms for the agents trained in the \pendulum~system without reward scaling. Shaded areas around the solid lines represent the $95\%$ confidence interval.
}
\label{fig:0_Pendulum_2000}
\end{figure*}

The result of training for $2000$ iterations without reward scaling is shown in Figure~\ref{fig:0_Pendulum_2000}. The average performance and number of arms at the end of training are recorded in Table~\ref{table:pendulum_results}. Both AQL and SPAQL distance themselves from the random policy, with SPAQL and SPAQL-TS performing better than AQL. The average number of arms of the AQL agents ($1.95\times10^{5}$) is two orders of magnitude higher than the average number of arms of the SPAQL and SPAQL-TS agents ($1.28\times10^{3}$ and $1.08\times10^{3}$, respectively), despite its lower performance. This exemplifies the problem of using time-variant policies to deal with time-invariant problems. 
Applying the Welch test \citep{colas2018many} to SPAQL and SPAQL-TS with a significance level of $5\%$, there is not enough evidence to support the claim that SPAQL is better than SPAQL-TS, and vice versa.

\begin{figure*}[t!]
\centering
\includegraphics[width=0.75\columnwidth]{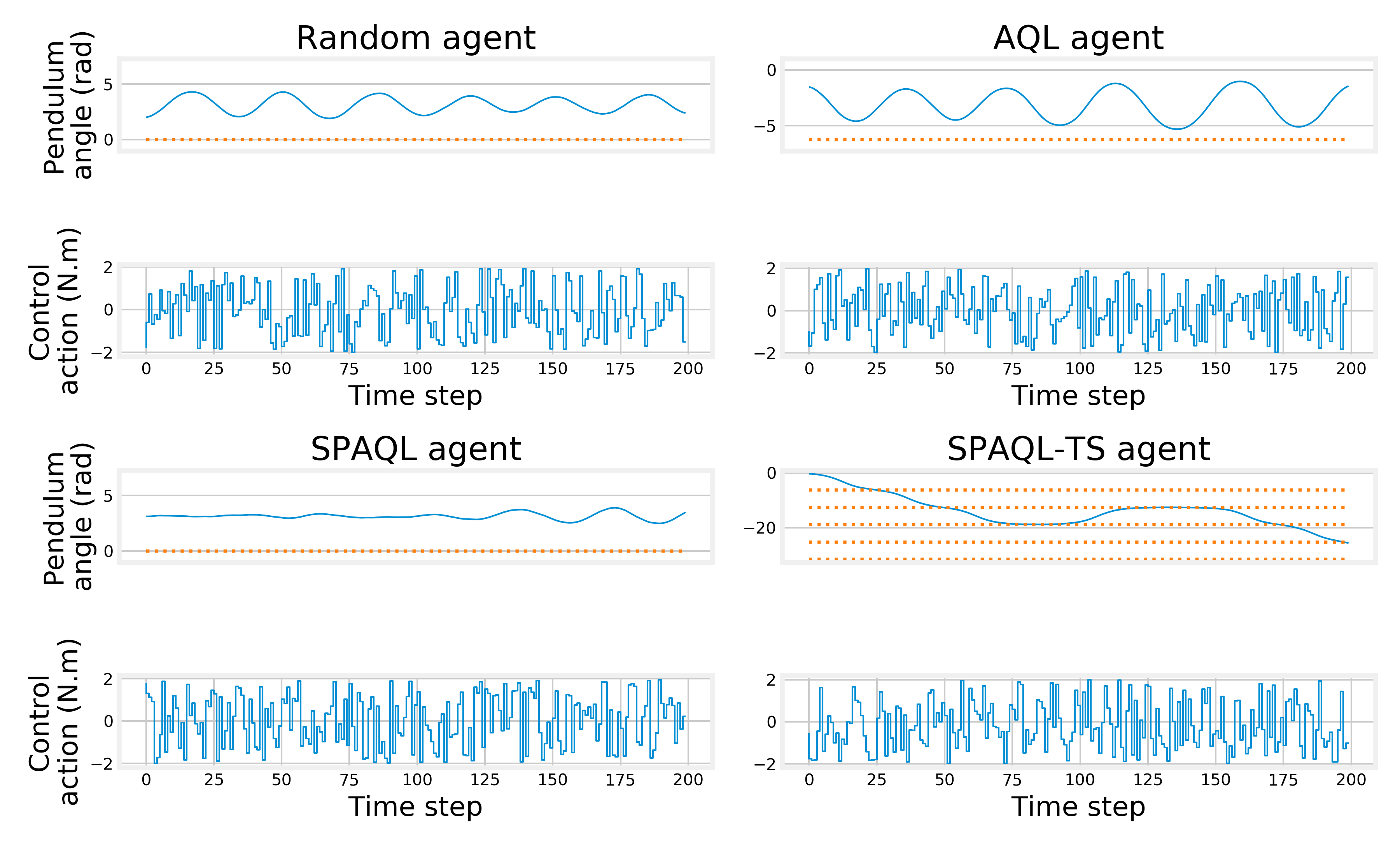}
\caption{Control action and evolution of the pendulum angle along a full episode for the four different policies being compared. The yellow dashed lines show the reference angle ($2\pi k$ radians, where $k$ is any integer).
}
\label{fig:0_Pendulum_2000_state_variables}
\end{figure*}

The evolution of the pendulum angle along a full episode under each of the policies after training is shown in Figure~\ref{fig:0_Pendulum_2000_state_variables}. Neither policy is able to stabilize at the reference angle (the SPAQL-TS agent manages to achieve temporary stability between time steps $75$ and $90$, and between $120$ and $150$). Looking at the control actions, they are all very similar to the random one. This can be attributed to the continuous action space.

\begin{figure*}[t!]
\centering
\includegraphics[width=0.75\columnwidth]{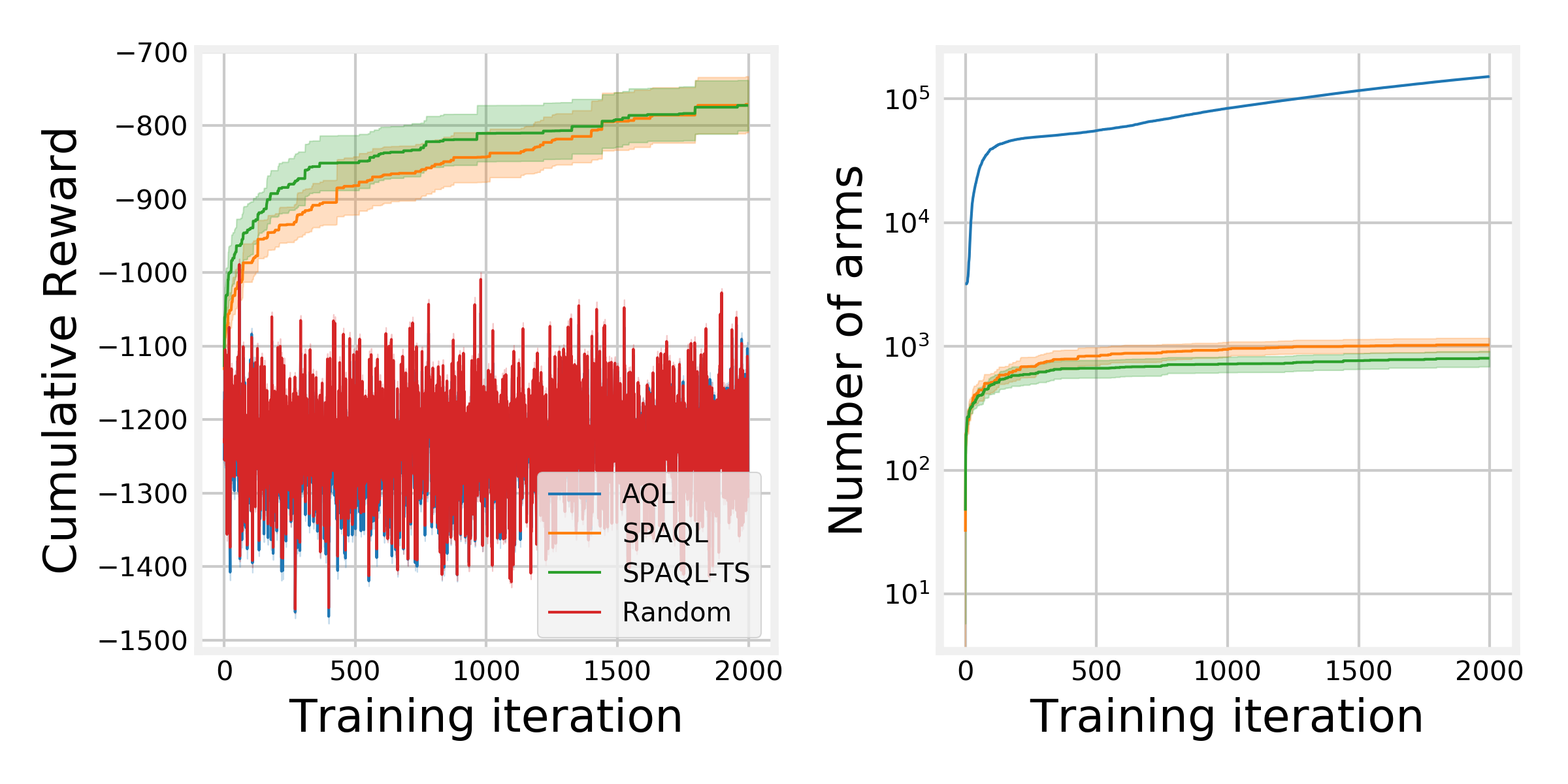}
\caption{Average cumulative rewards and number of arms for the agents trained in the \pendulum~system with a discrete action space. Shaded areas around the solid lines represent the $95\%$ confidence interval. Compared to Figure~\ref{fig:0_Pendulum_2000}, the cumulative reward achieved by SPAQL and SPAQL-TS at the end of training increased by about $100$.
}
\label{fig:0_PendulumCategorical}
\end{figure*}

\begin{figure*}[t!]
\centering
\includegraphics[width=0.75\columnwidth]{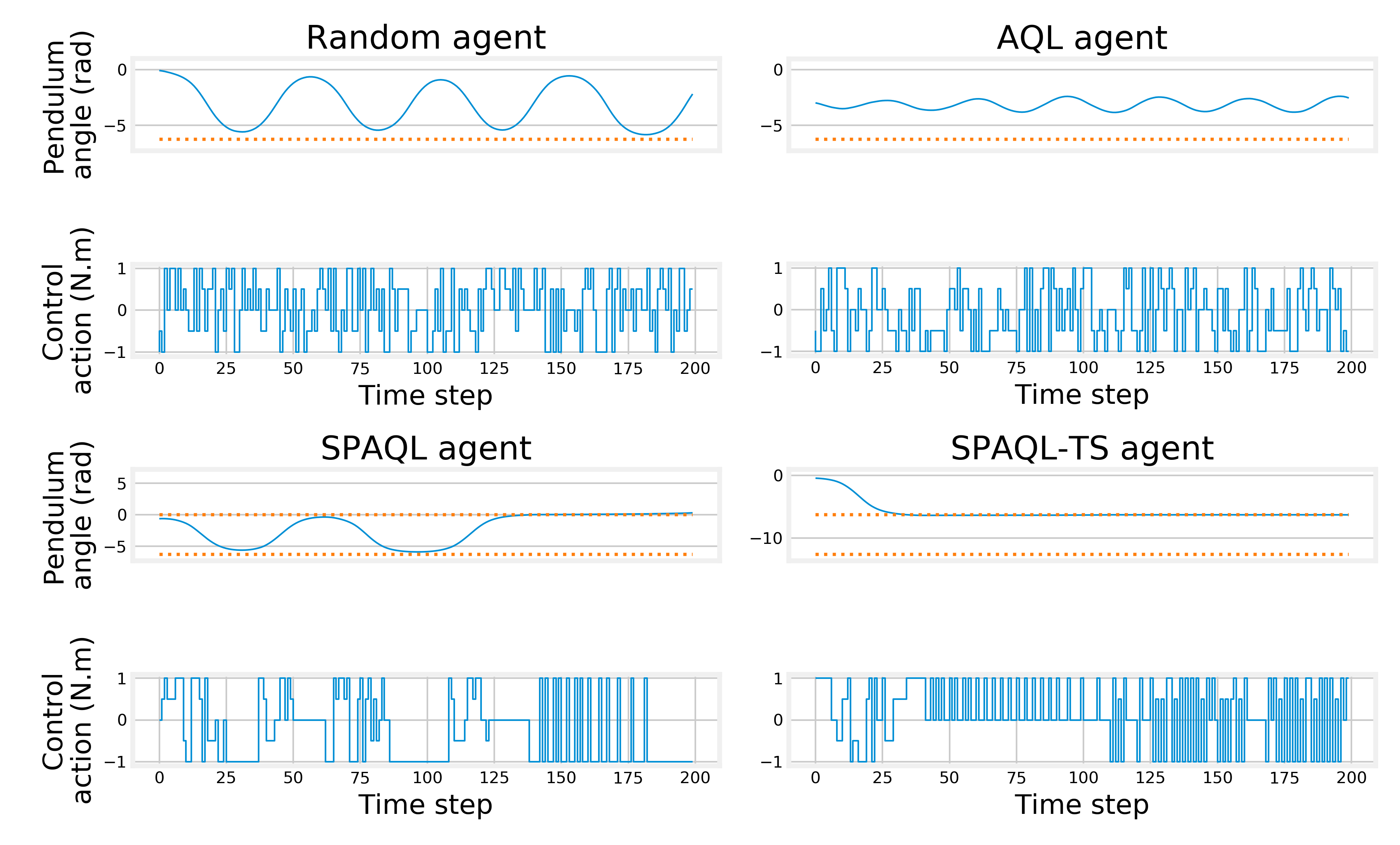}
\caption{Control action and evolution of the pendulum angle along a full episode for the four different policies being compared given that the action space has been limited to five actions. The yellow dashed lines show the reference angle ($2\pi k$ radians, where $k$ is any integer).
}
\label{fig:0_PendulumCategorical_state_variables}
\end{figure*}

Another experiment was run using a discrete action space ($5$ actions, $[-2, -1, 0, 1, 2]$). This leads to an increase in the proportion of episodes where the pendulum becomes controlled, as seen for example in Figure~\ref{fig:0_PendulumCategorical_state_variables}. This is accompanied by the increase in average cumulative reward seen in Figure~\ref{fig:0_PendulumCategorical}. It is worth noticing that using the simpler action space not only leads to the stable control solution faster, but it also results in policies that are easier to interpret.

\subsection{CartPole}

\subsubsection{Setup}
\label{section:experiments_set_up_cartpole}

The cartpole system consists of a pole with length $l = 1$m and mass $m = 0.1$kg attached to a cart of mass $m = 1$kg \citep{barto1983neuronlike}. The state vector considered is the cart's position $x$ and velocity $\dot{x}$, and the pole's angular position with the vertical axis $\theta$ and angular velocity $\dot{\theta}$. In order to distinguish the position $x$ from the state vector, in this section the state vector is denoted as $\mathbf{x} = [x, \dot{x}, \theta, \dot{\theta}]^T$. Unlike the \pendulum~system, in the \cartpole~system, the agent has direct access to the state variables. A simulation of the problem is termed an episode. An episode terminates when one of the following conditions is met \citep{1606.01540}: the episode length ($200$) is reached; the cart position leaves the interval $[-2.4, 2.4]$ meters; the pole angle leaves the interval $[-0.21, 0.21]$ rad ($[-12, 12]$ degrees).

The goal is to keep the simulation running for as long as possible (the $200$ time steps). Only two actions are allowed (``push left'' and ``push right'', with a force of $10$N). The cost $J$ is $-\sum_{t=1}^T 1$, where $T \leq H$ is the terminal time step \citep{1606.01540}. This is equivalent to a reward of $1$ for each time step, including the terminal one $T$. According to the \gym~documentation, the problem is considered solved when the average cost is lower than or equal to $-195$ (or the cumulative reward is greater than or equal to $195$) over $100$ consecutive trials. The concept of ``solved'' is used as an entry requirement for the \gym~Leaderboard, which serves as a tool for comparing RL algorithms \citep{1606.01540}.

The state-action space of this problem (as observed by the agent) is \citep{1606.01540}

\[
\S \times \A = \left(\left[-4.8, 4.8\right] \times \mathbb{R} \times \left[-\frac{24\pi}{180}, \frac{24\pi}{180}\right] \times \mathbb{R}\right) \times \{0,1\}.
\]
This space is mapped to a standard $[-1,1]^4\times\{0,1\}$ space. For the cart position $x$ and pole angle $\theta$ (first and third state variables, respectively), the mapping is done in a similar way as was done with the angular velocity of the \pendulum~(divide by the maximum value). Their range is twice the one mentioned in the termination conditions ($[-2.4, 2.4]$ meters and $[-12\pi/180, 12\pi/180]$ radians, respectively), in order to capture the terminal observation \citep{1606.01540}. Cart and pole velocities are allowed to assume any real value. To bound them, the sigmoid function (shifted and rescaled) is used

\[
\phi_m(y) = \frac{2}{1 + \exp \left(-2\,m\,y \right)} - 1.
\]
Parameter $m$ controls the slope at the origin, and can be used to include some domain knowledge. For example, the largest linear distance that the cart can travel without terminating an episode is $2.4 - (-2.4) = 4.8$m, and the time discretization step for this system is $0.02$. Therefore, the maximum absolute velocity which is reasonable to expect the cart to have is $4.8/0.02 = 240$m/s. This leads to $m_{\dot{x}} = 1/240$ as a reasonable value for $m$ when bounding the cart velocity. Following a similar reasoning, a reasonable value for $m$ when bounding the pole's angular velocity is $m_{\dot{\theta}} = 1/21$.

The $\infty$ product as metric for the \cartpole~problem is written as

\[
\D((\mathbf{x},u), (\mathbf{x}',u')) = \max\left\{\frac{|x - x'|}{4.8}, |\phi_{m_{\dot{x}}}(\dot{x}) - \phi_{m_{\dot{x}}}(\dot{x'})|, \frac{|\theta - \theta'|}{24\pi/180}, |\phi_{m_{\dot{\theta}}}(\dot{\theta}) - \phi_{m_{\dot{\theta}}}(\dot{\theta'})|, 2\times \Ind{u \neq u'}\right\},
\]
where $\Ind{A}$ is the indicator function ($1$ if condition A is true and $0$ otherwise). For SPAQL-TS, the reference state $\mathbf{x}_{ref}$ was set to $[0, 0, 0, 0]^T$.

This problem differs from the previous one in two important aspects. First, the rewards are already normalized. Second, while in the \pendulum~problem the set of possible actions is a real interval ($[-2, 2]$), in this problem the set of possible actions is finite (there are only two possible actions). In a certain sense, the \pendulum~problem is akin to a regression problem, while the \cartpole~is akin to a classification one. While in regression problems the objective is to learn a continuous mapping from inputs to outputs, in classification problems the objective is to learn a partition of an input space. Given the nature of AQL and SPAQL, it would be reasonable to assume that they would perform better in the \cartpole~problem than in the \pendulum~one. However, the \cartpole~problem also has a higher dimensional state-action space, which poses an additional challenge.

\subsubsection{Results}
\label{cartpole_experimental_results}

\begin{figure*}[t!]
\centering
\includegraphics[width=0.6\columnwidth]{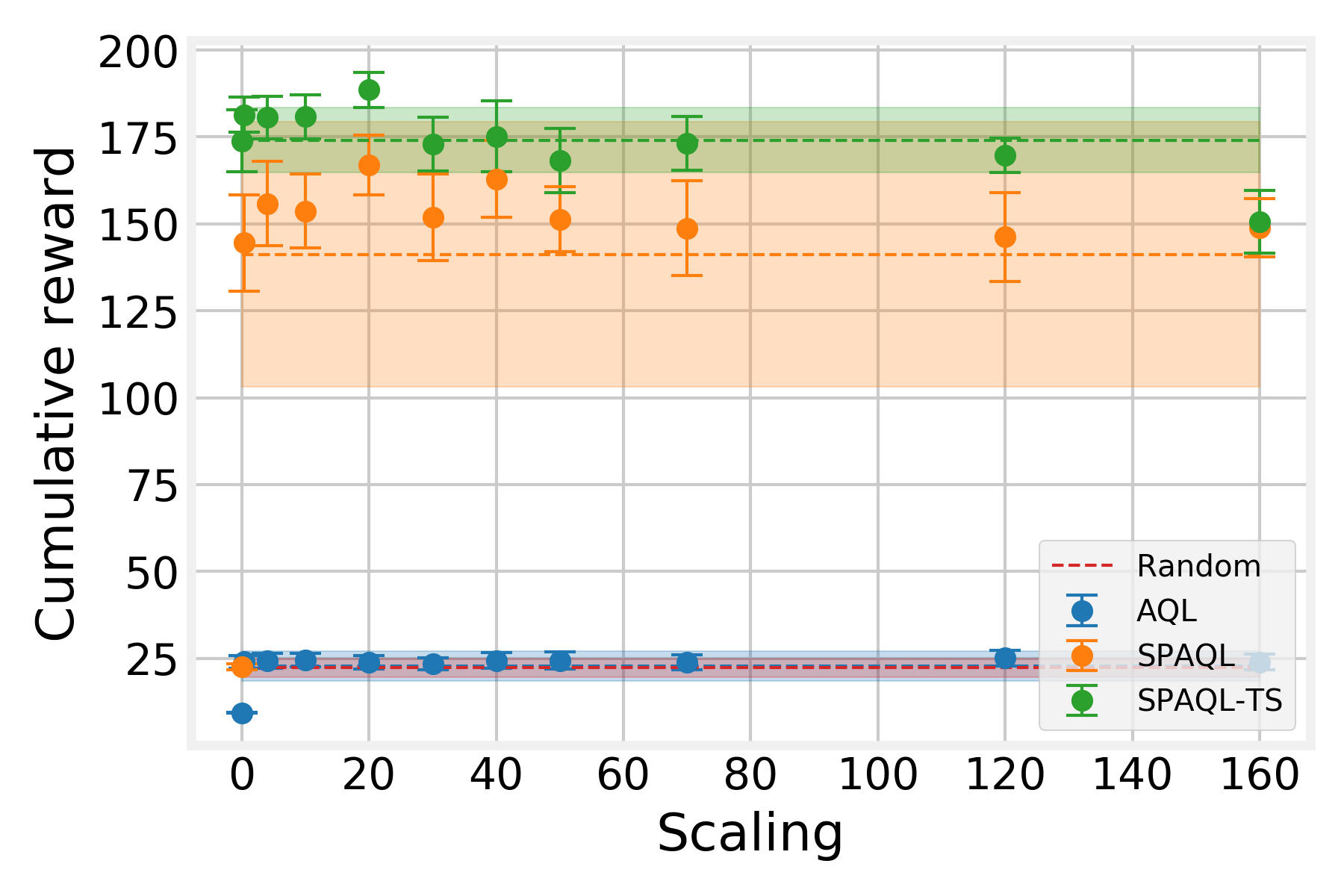}
\caption{
Comparison of different scaling parameter values on the average cumulative reward for the \cartpole~system. Each dot corresponds to the average of the rewards obtained by the $20$ agents after $100$ training iterations, and the error bars display the corresponding standard deviation. Dashed lines represent the average cumulative reward calculated over the scaling values listed in Section~\ref{sec:exp_proc_param}, and the shaded areas represent the corresponding standard deviation.
}
\label{fig:cartpole_scaling_results}
\end{figure*}

The effect of the scaling parameter $\xi$ on the \cartpole~problem is shown in Figure~\ref{fig:cartpole_scaling_results}, and leads to two observations. The first one is that disabling the upper confidence bounds (setting $\xi=0$) results in a great decrease in SPAQL performance, highlighted by the wide orange area. This negative effect is also seen in AQL. When $\xi = 0$, AQL performs worse than a random policy, and SPAQL performs as well as a random policy. Setting the scaling parameter to a relatively low value (such as $0.4$) immediately solves this problem. SPAQL-TS is not affected by this. Besides this first outlier ($\xi = 0$) in AQL and SPAQL, the effect of the scaling parameter on the three algorithms is not significant. The experimental evidence presented here supports the suggestion from \citep{Araujo20} of using non-zero scaling values lower than $H/3$.

The second observation regarding Figure~\ref{fig:cartpole_scaling_results} is that the AQL policies are barely distinguishable from random ones, independently of $\xi$. This may be due to the curse of dimensionality. With a $5$D state-action space, every time a split occurs, $32$ new balls are created\footnote{In fact, since there are only two possible actions, after the first split, the child balls will only be split into $16$ new ones.}. The updates to the values of the Q-function of these balls become independent. Updating one of those $32$ balls will lead to learning something related with that state. However, the remaining $31$ balls will remain random (equally likely to pick one of the two actions). This problem can be considered as a variant of the Coupon Collector Problem \citep{FLAJOLET1992207}. Assuming that the state-action space is visited uniformly at random, we wish to know how many visits are needed in order to be sure that each ball was visited at least once. This number is bounded\footnote{$f(n) = \Theta(g(n))$ means that $f$ is asymptotically bounded above and below by $g$.} by $\Theta(n \log (n))$, where $n$ is the number of balls in the partition. Given that $n$ grows exponentially with the dimension of the state-action space, the number of visits required will also grow exponentially. In AQL, we need to multiply this number of visits by $H$, since a different partition is kept for each time step, and they are updated independently. This argument seems to indicate that the practical sample efficiency of AQL does not scale well with the dimension of state-action space.

\begin{figure*}[th!]
\centering
\includegraphics[width=0.75\columnwidth]{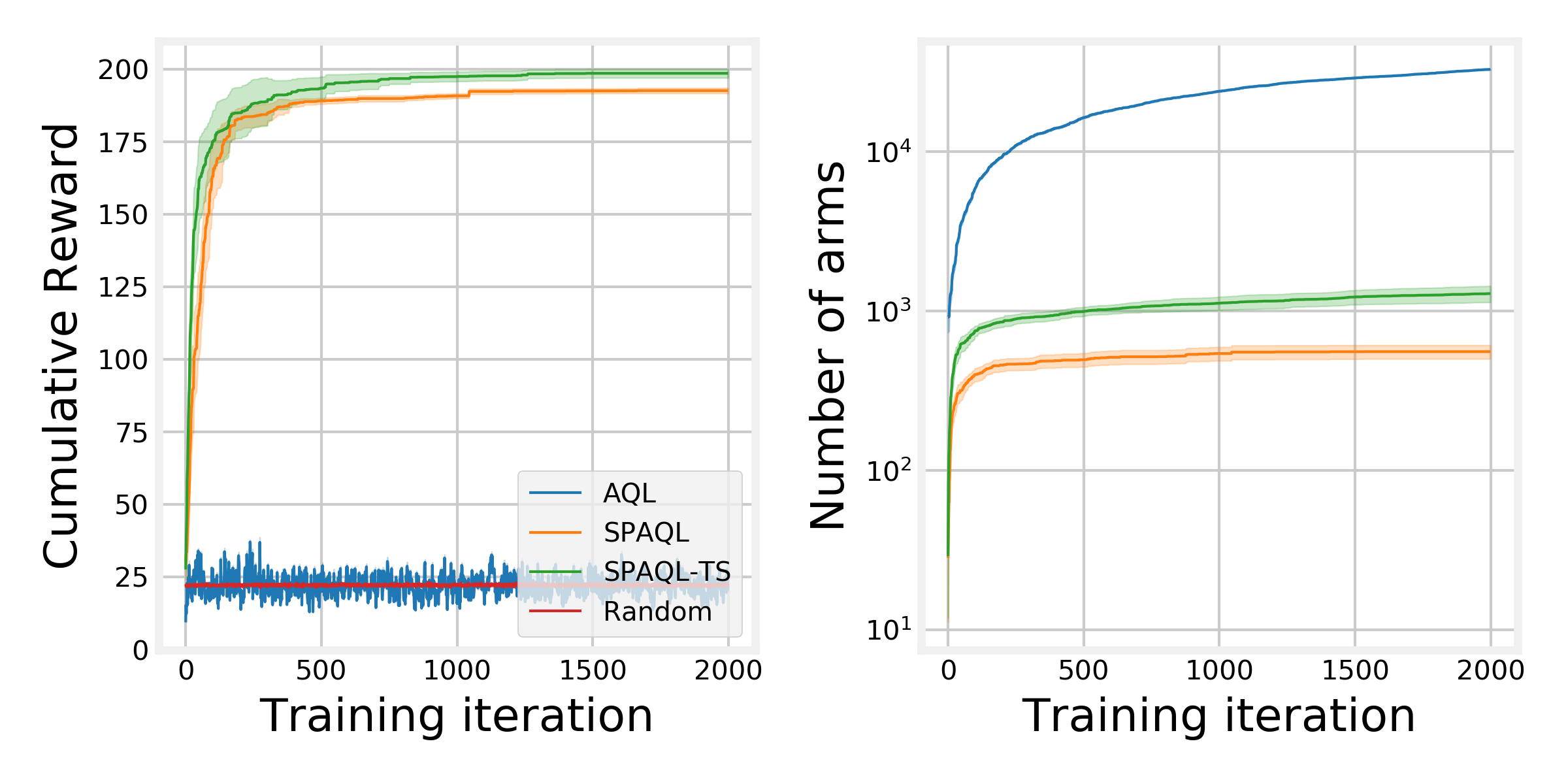}
\caption{Average cumulative rewards and number of arms for the agents trained in the \cartpole~system. At the end of each iteration the agents were evaluated $100$ times. Shaded areas around the solid lines represent the $95\%$ confidence interval.
}
\label{fig:CartPole_2000}
\end{figure*}

The training curves for the three algorithms are shown in Figure~\ref{fig:CartPole_2000}. The average performance and number of arms at the end of training are recorded in Table~\ref{table:cartpole_results}. While in the \pendulum~system the AQL agents were able to distinguish themselves from the random policy, for the \cartpole~system, this does not happen. There are fluctuations in the performance of AQL, although no improvement is permanent. On the other hand, the SPAQL and SPAQL-TS agents quickly achieve average cumulative rewards of approximately $193$ and $199$, respectively (out of $200.00$). Considering that these agents were evaluated $100$ times, the system is solved by SPAQL-TS. Manually inspecting the performances of the twenty individual agents stored at the end of training, it is seen that only one SPAQL agent solved the system (average cumulative reward higher than $195$). On the other hand, seventeen SPAQL-TS agents solved it (out of which fourteen scored the maximum average performance of $200$). The Welch test concludes that SPAQL-TS is better than SPAQL at significance level of $5\%$ with a p-value of the order of $10^{-6}$. SPAQL-TS agents end the training with around twice the number of arms of SPAQL ($1.29\times10
^{3}$ and $5.58\times10
^{2}$, respectively). The AQL agents finish training with $25$ times more arms ($3.30\times10
^{4}$).

\begin{figure*}[th!]
\centering
\includegraphics[width=0.75\columnwidth]{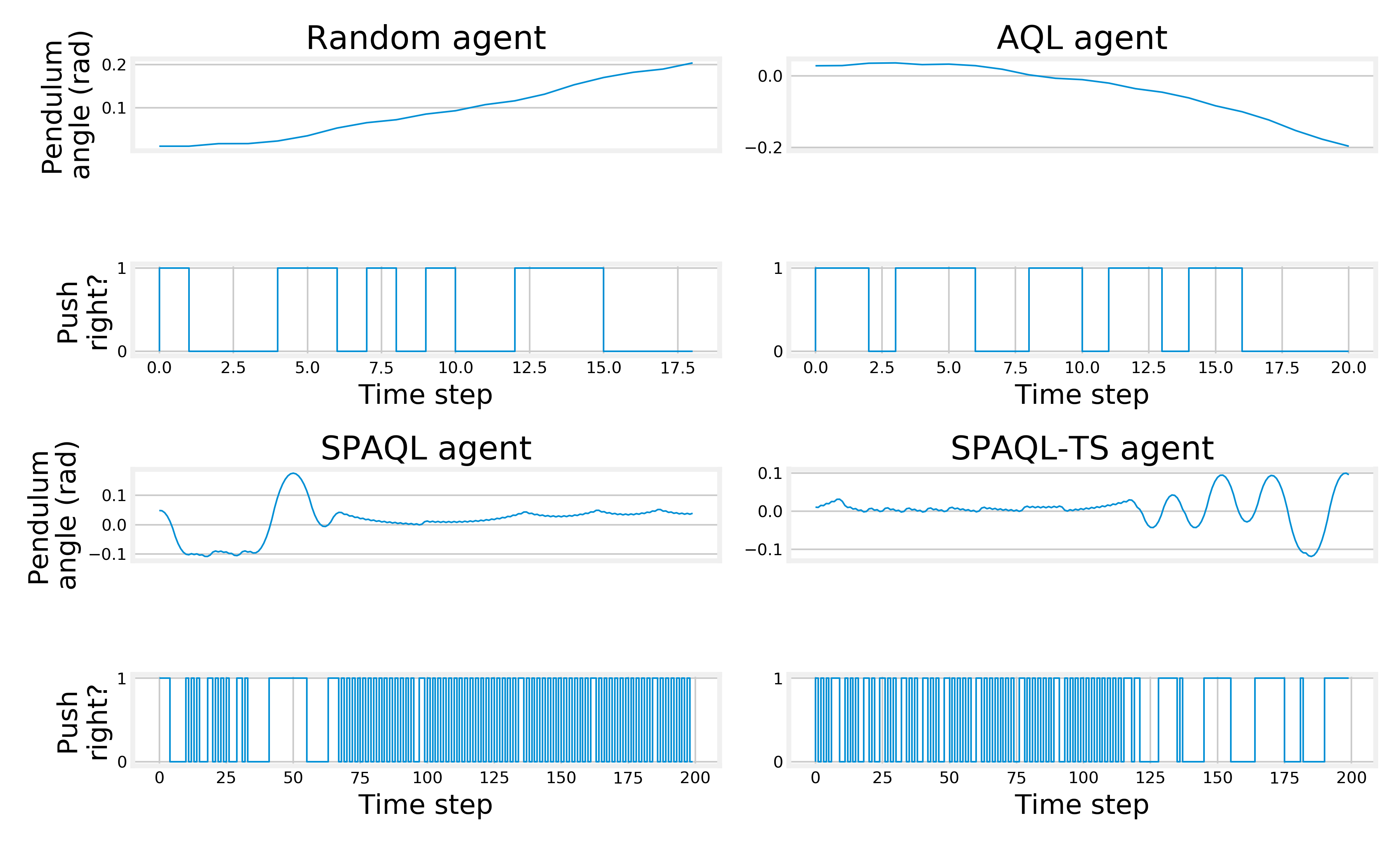}
\caption{Pole angle and control actions for the four different policies being compared.
}
\label{fig:CartPole_2000_state_variables}
\end{figure*}

The pole angle and the control actions are shown in Figure~\ref{fig:CartPole_2000_state_variables}. Both the random and AQL agents are unable to control the pole, and the episode terminates at most on time step $20$, when the pole angle reaches $0.2$ radians. For the SPAQL and SPAQL-TS agents, the pole angle is kept mostly between $-0.1$ and $0.1$ radians. Both episodes terminate at time step $200$. The behaviour seen towards the end of the SPAQL-TS episode shown in Figure~\ref{fig:CartPole_2000_state_variables} is an unavoidable consequence of only having two discrete control actions. However, this does not mean that, given more time steps, the pole angle would drift to the point where the episode would terminate. On the contrary, the previous results indicate that it is much more likely that the pendulum would be driven back to the vertical position.

\subsection{Comparing AQL, SPAQL, and TRPO}

\begin{table}[th!]
\centering
\begin{tabular}{lll}
         & \multicolumn{2}{c}{\pendulum} \\ \hline
         & \multicolumn{1}{c}{\begin{tabular}[c]{@{}c@{}}Average cumulative reward at\\ the end of training\end{tabular}} & \multicolumn{1}{c}{\begin{tabular}[c]{@{}c@{}}Number of arms at\\ the end of training\end{tabular}} \\ \hline
AQL      & $-1287.31 \pm 5.79$ & $195032.75 \pm 575.81$ \\
SPAQL    & $-835.99 \pm 26.91$ & $1277.50 \pm 168.74$ \\
SPAQL-TS & $-873.40 \pm 39.94$ & $1082.50 \pm 91.42$ \\\hline
Random   & $-1340.43 \pm 6.24$ &   \\
TRPO     & $\mathbf{-176.76 \pm 15.79}$ &   \\\hline                                        
\end{tabular}
\caption{Rewards and number of arms ($\pm 95\%$ confidence interval) for the different agents at the end of training in the \pendulum~system. The best performance is shown in bold.}
\label{table:pendulum_results}
\end{table}

\begin{table}[th!]
\centering
\begin{tabular}{lll}
         & \multicolumn{2}{c}{\cartpole} \\ \hline
         & \multicolumn{1}{c}{\begin{tabular}[c]{@{}c@{}}Average cumulative reward at\\ the end of training\end{tabular}} & \multicolumn{1}{c}{\begin{tabular}[c]{@{}c@{}}Number of arms at\\ the end of training\end{tabular}} \\ \hline
AQL      & $22.40 \pm 0.60$ & $32988.9 \pm 186.61$ \\
SPAQL    & $192.57 \pm 0.94$ & $557.75 \pm 55.10$ \\
SPAQL-TS & $\mathbf{198.53 \pm 1.55}$ & $1289.75 \pm 151.22$ \\\hline
Random   & $22.15 \pm 0.64$ &   \\
TRPO     & $\mathbf{197.27 \pm 3.17}$ &   \\\hline                                        
\end{tabular}
\caption{Rewards and number of arms ($\pm 95\%$ confidence interval) for the different agents at the end of training in the \cartpole~system. The best performance is shown in bold. When the Welch t-test does not find a significant statistical difference between two performances, both are shown in bold.}
\label{table:cartpole_results}
\end{table}

Trust region policy optimization (TRPO) is an on-policy method that is known to perform well in control problems \citep{trpo, duan2016benchmarking}. Instead of learning a value function and then choosing actions greedily, it learns a policy directly using neural networks. While this allows TRPO to perform well, and to generalize to unseen situations, the policies that it learns are not interpretable, unlike the ones learned by SPAQL (which are tables of actions associated with states). This trade-off between performance and interpretability justifies the comparison between these two algorithms.

\begin{figure*}[t!]
\centering
\includegraphics[width=0.8\columnwidth]{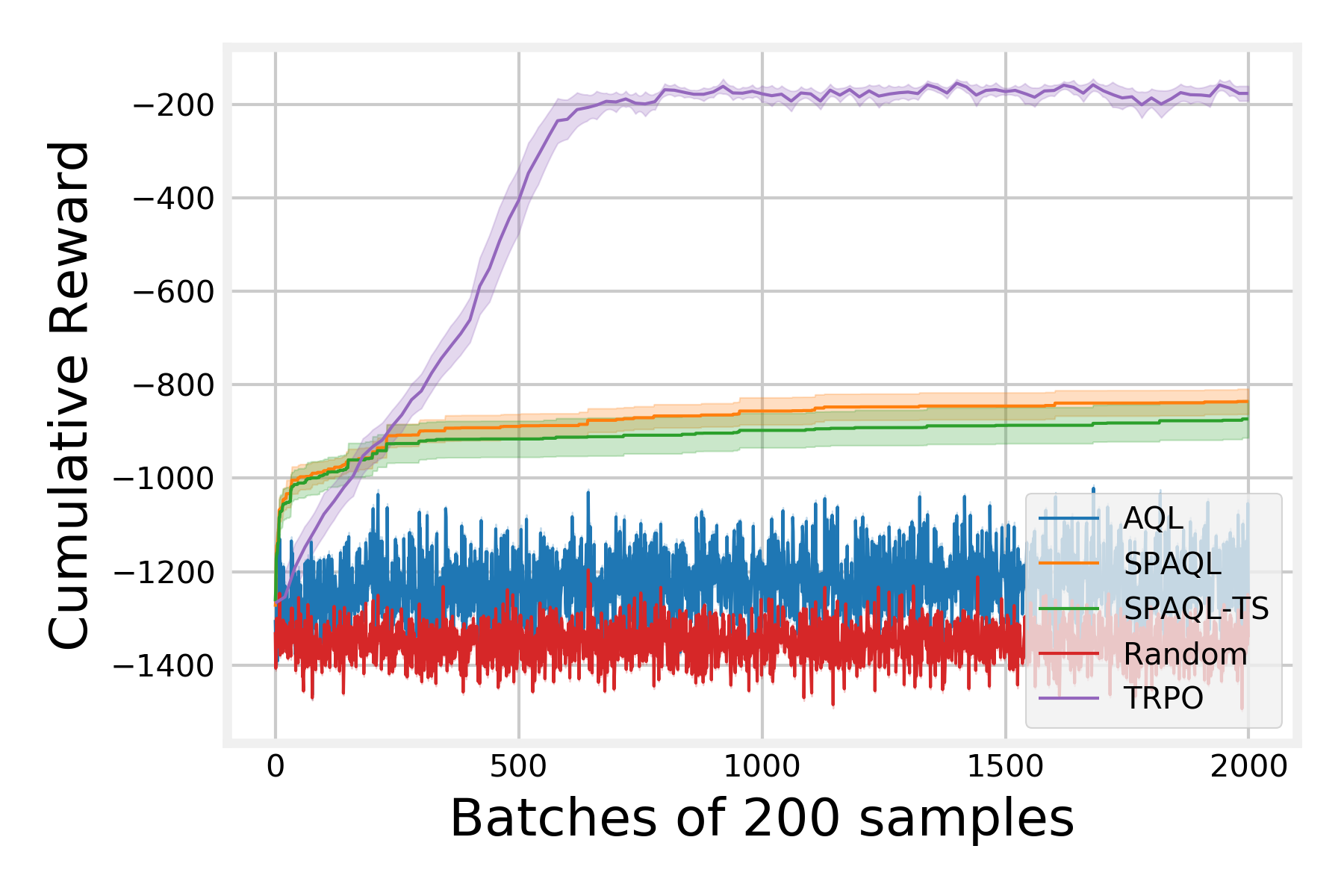}
\caption{Average cumulative rewards and number of arms for the agents trained in the \pendulum~system without reward scaling. Shaded areas around the solid lines represent the $95\%$ confidence interval.
}
\label{fig:Pendulum_2000_TRPO}
\end{figure*}

\begin{figure*}[th!]
\centering
\includegraphics[width=0.6\columnwidth]{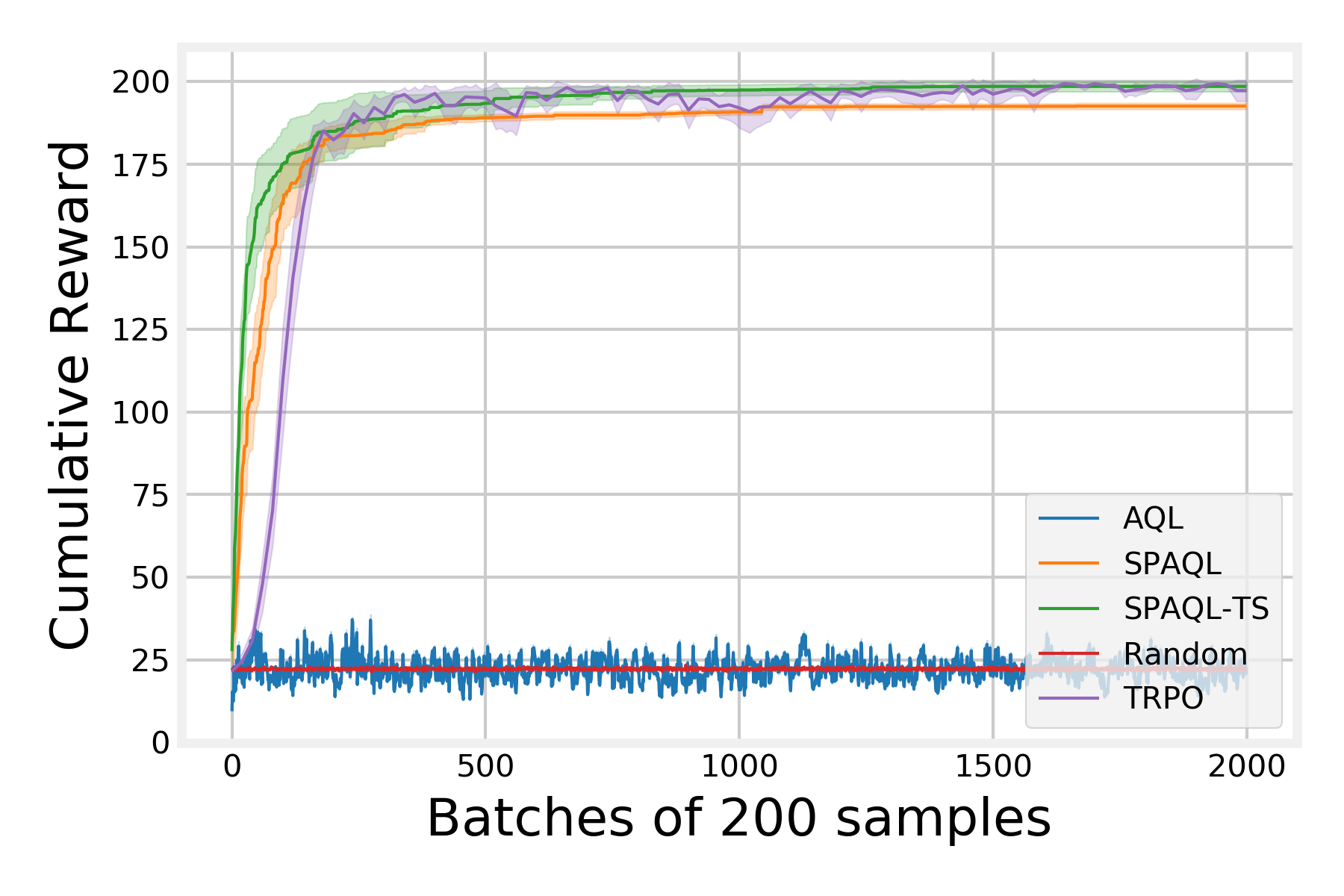}
\caption{Average cumulative rewards and number of arms for the agents trained in the \cartpole~system. Shaded areas around the solid lines represent the $95\%$ confidence interval.
}
\label{fig:CartPole_2000_TRPO}
\end{figure*}

The implementation of TRPO provided by \texttt{Garage} \citep{garage} was used. The parameters used in the examples provided with the source were kept. Average cumulative reward estimates were computed over $20$ different random seeds, with policies trained for $100$ iterations, using $4000$ samples per iteration. This gives a total of $400$ thousand training samples, the same amount used when training AQL and SPAQL agents ($2000$ training iterations, with $200$ samples collected in each). Figures~\ref{fig:Pendulum_2000_TRPO} and~\ref{fig:CartPole_2000_TRPO} show the learning curves for random, AQL, SPAQL, SPAQL-TS, and TRPO agents, as a function of the number of samples used (in batches of $200$ samples).

For the \pendulum~system (Figure~\ref{fig:Pendulum_2000_TRPO}, Table~\ref{table:pendulum_results}) the SPAQL variants take the lead in cumulative reward increase, but are outrun by TRPO after $200$ batches (the number of samples exceeds $40$ thousand). The TRPO agents continue learning until stabilizing the cumulative reward around $-200$.

In the \cartpole~system (Figure~\ref{fig:CartPole_2000_TRPO}, Table~\ref{table:cartpole_results}), both SPAQL variants use the initial batches more efficiently. TRPO catches up around batch $200$ ($40$ thousand samples). In the end, TRPO agents perform as well as SPAQL-TS agents (the Welch test does not find enough evidence to prove that one is better than the other).

TRPO uses a neural network to approximate the policy. Since neural networks are universal function approximators, it was expectable to see them perform better than AQL and SPAQL in the \pendulum~problem, which is similar to a regression problem. Trying to solve the \pendulum~using AQL or SPAQL is similar to solving a regression problem using a decision tree. Although possible, the resulting tree may be very large.

For the \cartpole~problem, which is similar to a classification problem, the neural network requires more samples than SPAQL to learn an adequate state-action space partition. However, even though it manages to learn this partition, it is not straightforward to convert the neural network to an interpretable representation such as a table. SPAQL policies are tables, and therefore the policy obtained at the end of training is already interpretable, should it be required.

\section{Conclusions}
\label{section:conclusion}
This paper evaluates AQL and SPAQL on the \gym~implementation of two classical control problems: the \pendulum~and the \cartpole. It also introduces SPAQL-TS, an improved version of SPAQL that borrows concepts from control theory.  Both SPAQL and SPAQL-TS algorithms outperform AQL, thus contributing with more empirical evidence towards the claim that SPAQL may have a higher sample-efficiency than AQL. SPAQL-TS manages to solve the \cartpole~problem, showing higher sample-efficiency than TRPO when processing the first batches of samples.

The \pendulum~system has a continuous action space, while the \cartpole~system has a finite one. The results in the \pendulum~show that, given their current status, neither SPAQL nor SPAQL-TS are good choices when controlling systems with a continuous action space. It is preferable to use them in control problems where there is a finite set of control actions, such as the \cartpole. One of the reasons for this is that updates to a ball do not affect its neighbors. Propagating updates to the neighboring balls could be a way to improve generalization of SPAQL policies. Another possible approach to get SPAQL to work on continuous action spaces is to partition only the state space, and for each state space part learn a linear mapping from states to actions. Another possible line of work is to combine the adaptive partitioning of SPAQL with other approaches that use interpretable nonlinear function approximators, like the symbolic regression methodology proposed by \citet{Kubalik19}.

Discretizing the action space seems to help in the \pendulum~environment. However, the improvement is not enough to reach the reward levels of a method like TRPO, that deals directly with the continuous action space. The sample efficiency of SPAQL is higher than that of TRPO during the first training iterations, but then saturates. Further research into understanding why this happens could uncover design flaws in SPAQL, and help guide the development of other sample efficient algorithms.

The empirical evaluation in this paper is not exhaustive. The \gym~implements many more control problems, with more challenging state-action spaces. A possible direction for further work is to test SPAQL and SPAQL-TS on more problems.
This would result in more interpretable policies for these problems, that could provide insights for future research in control. Another possible direction is to study the causes behind the higher sample-efficiency of SPAQL and SPAQL-TS when processing the first training batches, compared to TRPO (refer to Figures~\ref{fig:Pendulum_2000_TRPO} and~\ref{fig:CartPole_2000_TRPO}). This may help pave the road for more sample-efficient control algorithms.

Finally, although it is straightforward to convert the partitions learned by SPAQL into tables, it would be interesting to implement an automatic translator from partition to decision tree, which would enable a clear and simplified setting to analyze the policies.

    \section*{Acknowledgments}
    	We would like to thank CMA - \textit{Centro de Matemática e Aplicações, Faculdade de Ciências e Tecnologia, Universidade Nova de Lisboa}, for providing access to their computational facilities, which were used to carry the calculations presented in this paper.
    	
    	This research work is supported by Fundação para a Ciência e Tecnologia, through IDMEC, under LAETA, project UIDB/50022/2020.
    \newpage
    \bibliographystyle{apalike}
    {\bibliography{controlspaql}}

\newpage
\appendix

\section{Domain of a ball}
\label{sec:domain}

\begin{figure}[!ht]
    \centering
    \includegraphics[width=0.6\columnwidth,trim={0 3cm 0 3cm},clip]{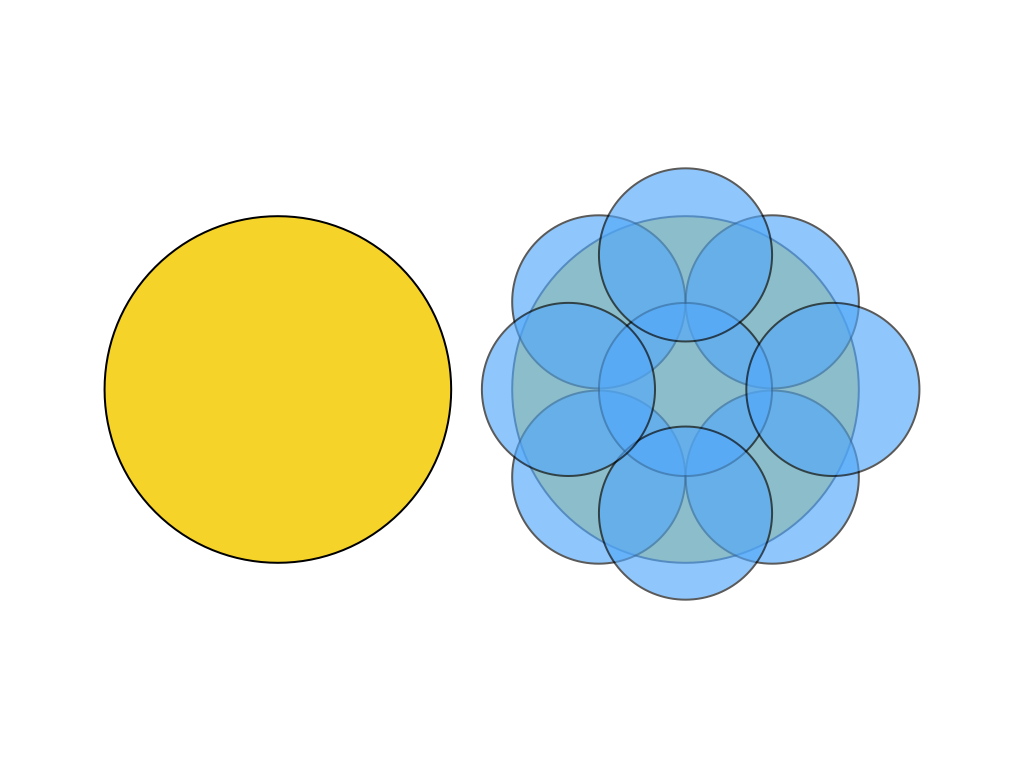}
    \caption{The initial ball (yellow ball on the left), and a blue covering of it (image on the right).}
    \label{fig:domain_1}
\end{figure}

\begin{figure}[!ht]
    \centering
    \includegraphics[width=0.6\columnwidth,trim={0 3cm 0 3cm},clip]{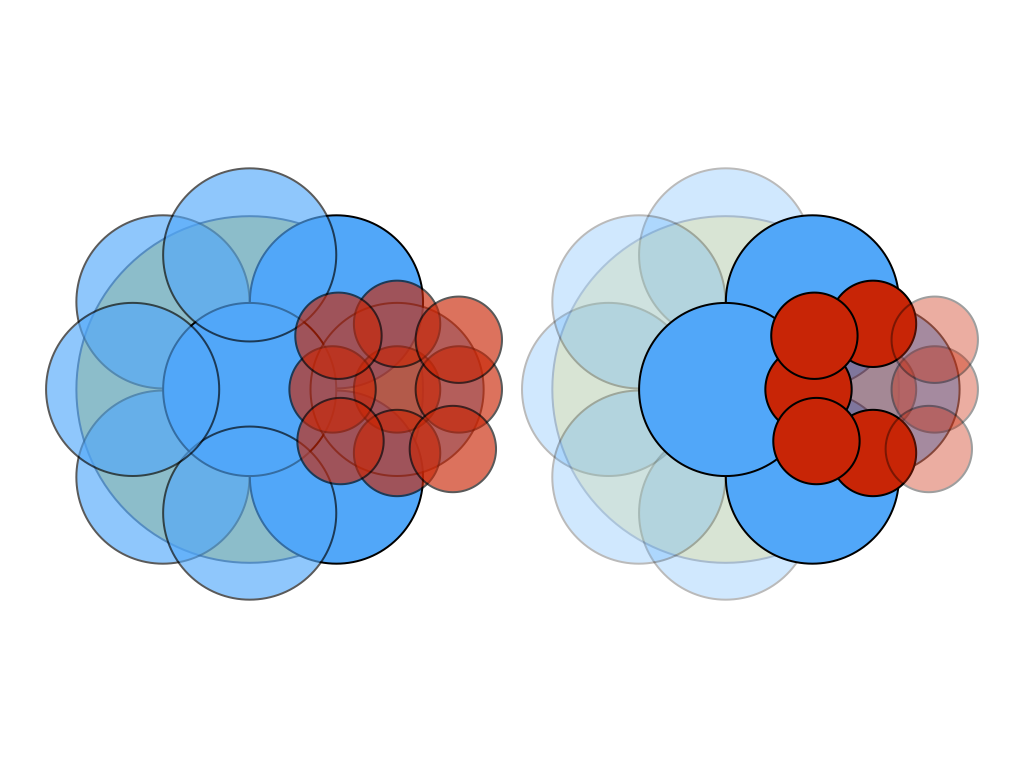}
    \caption{A covering of a blue ball (red balls on the left), and the domains of the neighboring blue balls (on the right).}
    \label{fig:domain_2}
\end{figure}

Consider the yellow ball in Figure~\ref{fig:domain_1} as the initial ball covering the entire state-action space. The blue balls on the right form a covering of the yellow ball. Since the radius of the blue balls is half of the radius of the yellow ball, and the blue balls cover the yellow ball completely, the domain of the yellow ball is the empty set. In Figure~\ref{fig:domain_2}, the red balls on the left image form a covering of the rightmost blue ball. Following a line of reasoning equal to the one used for the yellow ball, the domain of the blue ball covered by the red balls is the empty set. For the neighboring blue balls, that are not covered by the smaller red balls but intersect some of them, the domain is the set of points which are not contained inside any red ball. The domains are highlighted by the blue regions with full opacity seen on the image on the right. The points in the blue balls occluded by the red balls do not belong to the domain of the respective blue ball.

\newpage

\begin{figure}[!ht]
    \centering
    \includegraphics[width=0.6\columnwidth,trim={0 0cm 0 0cm},clip]{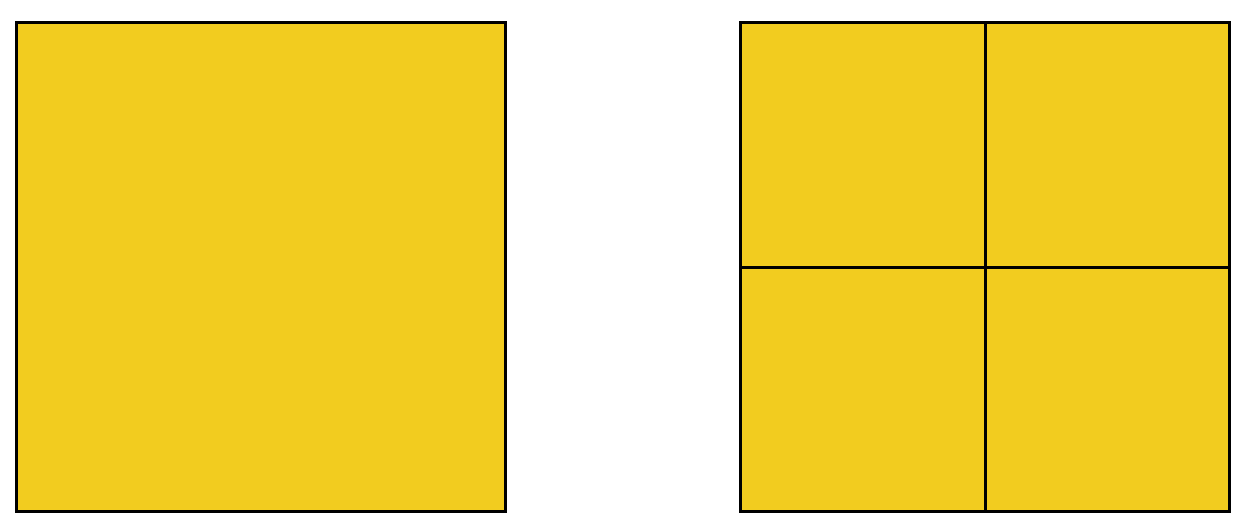}
    \caption{The initial ball (on the left), and a covering of it (on the right).}
    \label{fig:domain_3}
\end{figure}

Figure~\ref{fig:domain_3} illustrates a case where a disjoint covering can be found for every ball. The domain for each ball is either the ball itself (if it has never been split), or the empty set $\varnothing$ (if it has been split once).

\end{document}